\RequirePackage{amsthm}

\documentclass[pdflatex,sn-mathphys-num]{sn-jnl}

\usepackage{graphicx}%
\usepackage{multirow}%
\usepackage{amsmath,amssymb,amsfonts}%
\usepackage{amsthm}%
\usepackage{mathrsfs}%
\usepackage[title]{appendix}%
\usepackage{xcolor}%
\usepackage{textcomp}%
\usepackage{manyfoot}%
\usepackage{booktabs}%
\usepackage{enumitem}
\usepackage{algorithm}%
\usepackage{algorithmicx}%
\usepackage{algpseudocode}%
\usepackage{listings}%
\usepackage{lscape}
\usepackage{subcaption}
\usepackage{pdfpages}
\usepackage{makecell}
\usepackage{multirow}
\usepackage{amsmath}
\usepackage{xurl}
\usepackage{caption}  
\usepackage{array}   
\usepackage{lmodern}
\usepackage{anyfontsize}
\usepackage{placeins}  
\usepackage{needspace} 

\theoremstyle{thmstyleone}%
\theoremstyle{thmstyletwo}%

\theoremstyle{thmstylethree}%

\begin{document}

\title[Article Title]{Structured AI Decision-Making in Disaster Management}


\author*[1]{\fnm{Julian} \sur{Gerald Dcruz}}\email{juliangeralddcruz@outlook.com}

\author[1]{\fnm{Argyrios} \sur{Zolotas}}\email{a.zolotas@cranfield.ac.uk}

\author[2]{\fnm{Niall} \sur{Ross Greenwood}}\email{niall@neuron.world}

\author[3]{\fnm{Miguel} \sur{Arana-Catania}}\email{humd0244@ox.ac.uk}

\affil[1]{\orgdiv{Faculty of Engineering and Applied Sciences}, \orgname{Cranfield University}, \orgaddress{\country{UK}}}

\affil[2]{\orgdiv{Ageno School of Business}, \orgname{Golden Gate University}, \orgaddress{\country{USA}}}

\affil[3]{\orgdiv{Digital Scholarship at Oxford}, \orgname{University of Oxford}, \orgaddress{\country{UK}}}


\abstract{ 
With artificial intelligence (AI) being applied to bring autonomy to decision-making in safety-critical domains such as the ones typified in the aerospace and emergency-response services, there has been a call to address the ethical implications of structuring those decisions, so they remain reliable and justifiable when human lives are at stake.
This paper contributes to addressing the challenge of decision-making by proposing a structured decision-making framework as a foundational step towards responsible AI. The proposed structured decision-making framework is implemented in autonomous decision-making, specifically within disaster management. By introducing concepts of \textit{Enabler} agents, \textit{Levels} and \textit{Scenarios}, the proposed framework's performance is evaluated against systems relying solely on judgement-based insights, as well as human operators who have disaster experience: victims, volunteers, and stakeholders.
The results demonstrate that the structured decision-making framework achieves 60.94\% greater stability in consistently accurate decisions across multiple \textit{Scenarios}, compared to judgement-based systems. Moreover, the study shows that the proposed framework outperforms human operators with a 38.93\% higher accuracy across various \textit{Scenarios}. These findings demonstrate the promise of the structured decision‐making framework for building more reliable autonomous AI applications in safety‐critical contexts.
}




\maketitle

\section{Introduction}\label{sec1}
In disaster management, decision-making can be categorized into three phases: the pre-disaster phase, which involves sending warnings and planning evacuation procedures; the disaster phase, focused on rescue and relief operations; and the post-disaster phase, centred on assessing damage for reconstruction or rehabilitation activities \cite{RAUNGRATANAAMPORN2014658}. With the rapid advancement of artificial intelligence (AI), unmanned aerial vehicles (UAVs), and satellite imagery, significant efforts have been made to apply AI in providing judgment insights across each phase of the disaster management cycle \cite{make4020020}. While there have been initiatives to incorporate autonomous decision-making in tasks such as coordinating relief operations \cite{robertsEtAl15.flairs.coordinatingRobots}, these systems are scrutinized in safety-critical contexts due to the potential harm they may cause. This scrutiny has led to a paradox: {\em systems designed to enhance safety and efficiency are questioned for the risks they might introduce, particularly when errors or biases in autonomous decisions could have severe consequences}. This paradox underscores the tension between the advantages of automation and the need for meaningful human oversight to ensure accountability, fairness, and safety \cite{10.3389/fpos.2023.1238461}. To address the challenges of reliable and justifiable decisions in designing autonomous decision-making systems for safety-critical domains like disaster management, it is crucial to incorporate structure into AI decision-making. 

The work in this paper stems from a research project that proposed and designed a structured decision-making framework for autonomous decision-making in safety-critical domains. The work particularly contributes the following:
\begin{itemize}
    \item Designing and developing a structured decision-making framework for autonomous decision-making within disaster management.
    \item Developing an autonomous decision-making agent that makes use of the previous framework to enhance its decisions in disaster management. 
    \item Conducting a human evaluation study to effectively evaluate the autonomous decision-maker with human operators. 
\end{itemize}

The project's codebase is publicly available in the project's repository\\(\url{https://github.com/From-Governance-To-Autonomous-Robots/Autonomous-Governance-in-Disaster-Management}).\\

The structure of the paper is as follows: Section \ref{sec2} provides a brief survey of related work in the current literature. Section \ref{sec3} describes the proposed framework, followed by a detailed methodology of the approach in Section \ref{sec4}. Section \ref{sec5} presents and analyzes the experimental results, including a comparative analysis between human evaluation and the RL decision maker. Finally, Section \ref{sec6} draws conclusions from the study.

\section{Related Work}\label{sec2}
In disaster management, decision-making can be inherently complex, often unstructured, dynamic, and unpredictable. This process demands significant judgment and insight, frequently involving coordination between multiple local and government agencies, such as volunteer organizations and governmental authorities. Such unstructured decision-making can lead to unequal or delayed responses to victims' needs, as decisions are frequently based on immediate perceptions rather than structured data. For instance, the response to the Fukushima disaster highlighted the challenges faced by rescue teams in making critical decisions under uncertainty \cite{Arifah2019DECISIONMI}. Similarly, during a major railway accident in the UK, poor coordination among agencies significantly hampered the disaster response efforts \cite{Smith2000ACS}. 

Over the years, the application of machine learning in disaster management has significantly improved decision-making processes, particularly by enabling faster response times. The advent of surveillance drones, satellite imagery, and various data modalities has inundated decision makers with extensive data, necessitating the use of machine learning algorithms to generate explainable insights.

Previous research has explored various machine learning techniques to provide explainable insights during the disaster response phase, thereby supporting emergency response activities \cite{Amiran2024,Shukla2023,Sun2022,Fan2024}. In this context, \cite{Gupta2020AHM} proposed a framework utilizing satellite imagery from six types of natural disasters to identify the primary causes of damage in affected areas, achieving 99.59\% accuracy and facilitating effective interventions. Similarly,  \cite{FerdaOfli2020} introduced a multimodal deep learning model that integrates text and image data from social media, outperforming single-modality models. These studies demonstrated the viability of leveraging social media posts to triage victim requests and support decision-making in emergency response. Additionally, machine learning techniques have proven valuable in aiding decisions related to reconstruction and rehabilitation during the post-disaster phase. \cite{Yang2019} trained a CNN model combined with an ensemble max-voting classifier to assess flood damage in Texas using aerial images, achieving 85.6\% accuracy and an F1-score of 89.06\%. \cite{Dotel2020} developed a CNN model to assess the impact of water-related disasters by segmenting topographical features in pre- and post-disaster satellite images, identifying regions with significant changes. These studies demonstrated that machine learning techniques can expedite the analysis of satellite and drone data, reducing reliance on expert opinion and mitigating delays or inaccuracies in post-disaster decision-making. Inspired by these works, this paper utilizes the CrisisMMD Multimodal Twitter dataset \cite{FirojAlam2018}, an image-text dataset collected during various natural disasters, to address decision-making in the disaster phase of the disaster management cycle. CrisisMMD is the de-facto benchmark for image–text fusion during fast-moving natural hazards and contains 16\,058 tweets paired with 18\,082 images spanning seven 2017 disasters. Although modest in size, CrisisMMD remains the only openly licensed dataset in which every tweet–image pair is simultaneously annotated for (i) informativeness, (ii) humanitarian category, and (iii) damage severity. That unique tri-label design explains why virtually every multimodal disaster-response study since 2018 tests on it. The current state-of-the-art—transformer models with cross-modal attention and self-attention (e.g., \cite{Rezk2023MCA})—pushes the macro-F\textsubscript{1} ceiling to roughly 92 \%. Even so, performance is capped by two structural flaws: (1) noisy crowdsourced labels, and (2) temporal homogeneity—every event in the corpus happened in 2017. Minority classes such as missing people and vehicle damage remain chronically under-represented; most authors collapse them into broader buckets, a stop-gap that is anything but ideal for real-world triage. Additionally, the xBD dataset \cite{RitwikGupta2019}, containing pre- and post-event satellite imagery for various disasters, and the RescueNet dataset \cite{Rahnemoonfar_2023}, comprising post-disaster UAV images from multiple impacted regions, are utilized to address decision-making in the post-disaster phase. The xBD dataset is the largest fully-annotated satellite dataset for post-disaster building assessment: 22\,068 image pairs, 850\,736 building polygons, four-level “Joint Damage Scale”, and 80/10/10 train–test–holdout splits over 19 events and 45k km\textsuperscript{2} of ground coverage. It underpins the xView2 Challenge and therefore anchors most modern work on large-scale change detection.  Gupta et al.’s baseline CNN hits 0.59 macro F1; hierarchical transformer architectures \cite{Kaur2022DAHiTrA} now exceed 0.71 by explicitly modelling multi-resolution context and temporal deltas.  Yet xBD is still EO-only, dominated by U.S. and Western Pacific imagery, and heavily skewed toward the “no-damage” class (8× the next largest), so models trained here rarely transfer cleanly to SAR imagery or under-represented geographies. The RescueNet dataset fills the altitude gap with 4\,494 4096\,$\times$\,3072 UAV frames captured after Hurricane Michael, labelled at pixel-level for 11 semantic classes (from “road-blocked” to four granular building-damage states) and split 80/10/10 for train/val/test.  Baselines on DeepLabv3 + ResNet-101 reach 81\% mean IoU \cite{Rahnemoonfar_2023}, but performance collapses on the minority “road-blocked” and “pool” classes—evidence that high-resolution alone does not solve class imbalance.

However, these studies have not succeeded in addressing the challenges in the operations of disaster management which involves different agencies, including local governments, non-governmental organizations, and international bodies, and these agencies often follow varied protocols and communication standards, which leads to inter-agency coordination challenges. This lack of coherence can result in duplicated efforts, resource misallocation, and delays in critical response activities. During Hurricane Katrina, for example, misalignment between federal, state, and local authorities led to substantial delays in providing aid and evacuating affected individuals \cite{Bharosa2010Challenges}. Furthermore, current disaster management practices often struggle with information overload and the integration of diverse data sources. Stakeholders must frequently make rapid decisions based on incomplete or rapidly changing information, which can lead to potential oversights and mistakes \cite{MISRA2020101762}. Additionally, prolonged exposure to disaster-related decision-making can result in decreased accuracy and effectiveness among stakeholders, a phenomenon referred to as community disaster fatigue, particularly manifesting as a sense of defeatism \cite{INGHAM2023103831}. In response to these challenges, recent years have seen significant advancements in the introduction of AI for autonomous decision-making in disaster management. \cite{robertsEtAl15.flairs.coordinatingRobots} explored the coordination of semi-autonomous robot teams for disaster relief, focusing on the Situated Decision Process for managing tasks under uncertainty. They identified critical challenges in integrating reactive and deliberative planning, ensuring that autonomous vehicles execute their missions as intended without direct human oversight. Furthermore, \cite{Shah2019} examined the integration of big data analytics with IoT in disaster management, highlighting the difficulties in managing large volumes of data, making timely decisions, and maintaining the reliability of autonomous systems under crisis conditions. 

Most of these works focus on replacing the human in the loop for safety-critical decisions in disaster management. \cite{Nussbaumer2021} underscored the ethical implications of AI in safety-critical environments, particularly in disaster response contexts, and discussed the necessity of governance frameworks to ensure responsible, transparent, and fair decision-making processes. Since machine learning (ML) decisions are heavily data-driven and inherently unpredictable, these systems may exhibit significantly different behaviours in response to nearly identical inputs, making it difficult to define `correct' behaviours and ensure safety in advance \cite{Koopman2016}. Moreover, scholars have identified potential safety risks arising from the interaction between AI systems and their users, particularly due to automation bias. This bias occurs when humans attribute greater credibility to automated decisions because of their perceived objectivity, leading to complacency and less cautious behaviour when interacting with AI systems \cite{osoba2017intelligence,Taeihagh2019}. Autonomous decision-making significantly reduces human control, particularly in safety-critical decisions where AI systems are responsible for actions that impact human lives. Current legal frameworks typically place the responsibility for decision-making on humans, treating machine learning operations as tools that assist in the decision-making process \cite{matthias2004responsibility,leenes2014laws}. However, since AI largely depends on processes that independently develop and modify their own rules, human oversight is diminished. This reduction in control makes it unreasonable to hold human stakeholders or manufacturers fully accountable for the AI's actions, especially given the unpredictable nature of ML decisions, which implies that erroneous decisions made by AI are beyond the control and anticipation of both parties \cite{Butcher2019,kim2017crashed}.

The deployment of robots for personal care services, for instance, has sparked significant worries about the potential compromise of patient autonomy and dignity, particularly when robots impose strict limitations on patient mobility to prevent dangerous situations, see \cite{Leenes2017,TAN2021120686}. Similarly, the introduction of autonomous weapon systems, including drones and unmanned aerial vehicles, aimed at enhancing the precision and effectiveness of military operations, has raised concerns about the ethical and legal implications of such technologies \cite{Lele2019,HeatherRoff2014}. The delegation of authority to machines in making decisions about the use of lethal force, coupled with the lack of control over these autonomous systems, has also prompted significant ethical debates \cite{Firlej2021,scharre2016autonomous,solovyeva2018going}. 

To address these ethical implications, scholars have been proposing governance frameworks for responsible AI \cite{Papagiannidis2023,Ghallab2019,DIAZRODRIGUEZ2023101896}. \cite{Khanna_Khanna_Srivastava_Pandey_2021} proposed a governance framework for AI in oncology, where the decision-making on safety-critical decisions directly relates to the decision-making context of disaster management. The framework proposed by \cite{Khanna_Khanna_Srivastava_Pandey_2021} emphasized ethical soundness, legal compliance, equity, and patient-centred care, with an emphasis on quality assurance, transparency, bias mitigation, stakeholder engagement, and continuous adaptation through regular audits and feedback loops.


\section{Framework for Structured Decision Making}\label{sec3}
This section presents the proposed framework for structured decision-making. The framework aims to address inter-agency coordination and data overloading challenges, and mitigate the defeatism often experienced by stakeholders in disaster scenarios. 
The framework specifically focuses on implementing a structured decision-making approach, by clearly defining roles for the agents (\verb|Enabler| and \verb|Decision Maker| agents) and how they interact to make decisions at each \verb|Level| of the disaster management \verb|Scenario|. The framework aims to optimize decision-making flow through traceability of decisions in emergency response, rehabilitation, and reconstruction activities.


\begin{figure}[htp]
    \centering
    \includegraphics[width=0.9\linewidth]{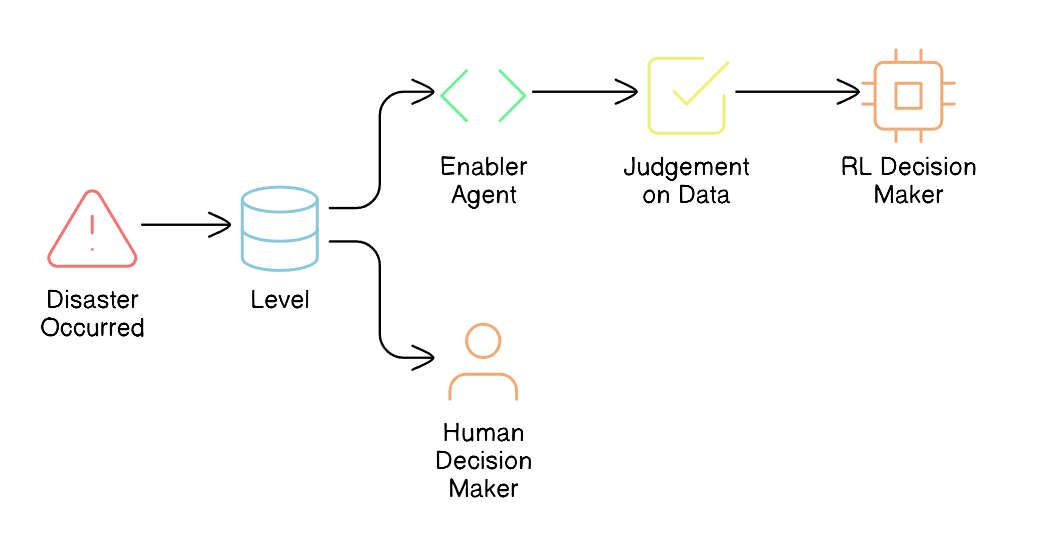}
    \caption{High-level overview of the proposed framework for structured decision-making in an autonomous system, compared with the decision flow of a human decision-maker.}
    \label{fig:systemDecision}
\end{figure}

\begin{figure}[htp]
\centering
\includegraphics[width=\textwidth,height=1.3\textheight,keepaspectratio]{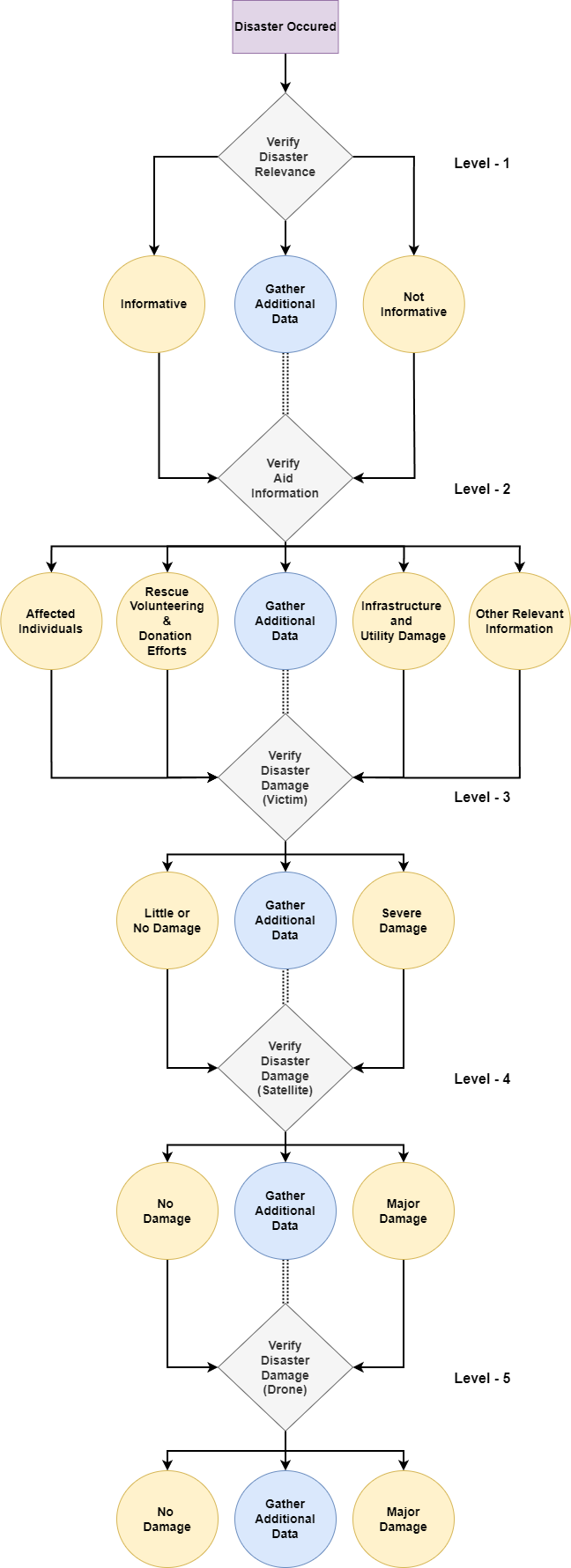}
\caption{System design of the proposed framework for structured decision-making in disaster management.}
\label{fig:systemDiagram}
\end{figure}

From a high-level perspective, the proposed framework for structured decision-making in disaster management comprises \verb|Enabler| agents and \verb|Decision Maker| agents, which collaborate to effectively manage disaster-phase and post-disaster phase scenarios. A representation of this framework is shown in Figure \ref{fig:systemDecision}, which also compares the decision flow with that of a human decision-maker in disaster management. To implement structured decision-making, the decision flow is organized into distinct \verb|Levels|, as illustrated in Figure \ref{fig:systemDiagram}. This structured approach is visually represented as a tree-like structure, referred to as a \verb|Scenario|. A \verb|Scenario| consists of five nodes, depicted as grey-filled boxes in the figure. Each node, known as a \verb|Level| within the framework, represents a stage where a critical decision must be made, potentially leading to a consequence or reward. The yellow-filled ellipses indicate the decision options available to the \verb|Decision Maker| agent based on the data received at that \verb|Level|. At each \verb|Level|, the data is first processed by an \verb|Enabler| agent before the \verb|Decision Maker| can make a decision. The \verb|Enabler| agent is an AI model trained to evaluate disaster-related data at that \verb|Level|, providing judgment insights (an array of confidence scores for each decision option) to assist the \verb|Decision Maker| in making informed decisions within a \verb|Scenario|. The \verb|Decision Maker| agent can be either a reinforcement learning (RL) algorithm or a human operator. However, if the \verb|Decision Maker| is a human operator, the \verb|Enabler| agent is not involved, as the human relies on their expertise and training to make decisions at each \verb|Level|. Additionally, the blue-filled ellipses in the diagram represent the \verb|Gather Additional Data| option, which serves as a real-world interaction point for the RL agent or human operator. This option allows for the acquisition of additional data at a given \verb|Level|.

\clearpage

One \verb|Scenario| comprises of the following levels:
\begin{itemize}
    \item \verb|Level|-1, where the data received at this stage is either informative or not informative to the disaster.
    \item \verb|Level|-2, where the data received at this stage is related to a type of humanitarian effort.
    \item \verb|Level|-3, where the data received at this stage is related to damage severity and was collected by victims or volunteers.
    \item \verb|Level|-4, where the data received at this stage is related to damage severity and was collected by satellite imagery.
    \item \verb|Level|-5, where the data received at this stage is related to damage severity and was collected by Unmanned Aerial Vehicles (UAVs).
\end{itemize}

\section{Methods}\label{sec4}
This section details the methodology followed in the implementation and evaluation of the proposed framework. The framework involves \verb|Enabler| agents and \verb|Decision Maker| agents collaborating across multiple \verb|Levels| in a tree-like \verb|Scenario| structure. Each \verb|Level| represents a critical decision point, where \verb|Enabler| agents process disaster-related data to assist the \verb|Decision Maker|, which can be either a reinforcement learning (RL) algorithm or a human operator. This section details each \verb|Level|, the training methodology of the RL agent, the evaluation metrics used, and the logic behind the web application designed for the evaluation, as well as the data used.

\subsection{Dataset Overview}\label{subsec3.1}
Our implementation focuses on decision-making during the disaster phase and post-disaster phase in disaster management. 
For disaster-phase decisions related to emergency response activities such as search, rescue, and relief operations, the CrisisMMD dataset \cite{FirojAlam2018} (a multimodal image-text dataset on disaster response activities collected from Twitter) is employed. CrisisMMD contains 16\,058 tweets paired with 18\,082 images spanning seven 2017 disasters. Every tweet–image pair is simultaneously annotated for (i) informativeness, (ii) humanitarian category, and (iii) damage severity. For post-disaster phase decisions involving damage assessment for reconstruction or rehabilitation activities, the xBD dataset (a dataset of images collected from satellites on pre- and post-disaster damage) \cite{RitwikGupta2019} and the RescueNet dataset (a dataset of images collected from drones on pre- and post- disaster damage) \cite{Rahnemoonfar_2023} are used.
The xBD dataset contains 22\,068 image pairs, 850\,736 building polygons, four-level “Joint Damage Scale”, and 80/10/10 train–test–holdout splits over 19 events and 45k km\textsuperscript{2} of ground coverage. xBD is still EO-only, dominated by U.S. and Western Pacific imagery, and heavily skewed toward the “no-damage” class (8× the next largest). The RescueNet dataset fills the altitude gap with 4\,494 4096\,$\times$\,3072 UAV frames captured after Hurricane Michael, labelled at pixel-level for 11 semantic classes (from “road-blocked” to four granular building-damage states) and split 80/10/10 for train/val/test. 

These datasets are used to train the \verb|Enabler| agents (which is discussed in more detail in the next section) that make up the structured decision-making. Each dataset will be presented in detail in the following sections together with each of the machine learning models using it.


\subsection{Enabler Agent}
The \verb|Enabler| agent supports the RL \verb|Decision Maker| agent by providing judgment insights on the data received at each \verb|Level| of a \verb|Scenario|, enabling informed decision-making. This section outlines the methodology used to train the \verb|Enabler| agents for the \verb|Levels| corresponding to the disaster-phase and post-disaster phase of the disaster management cycle.

\subsubsection{Disaster Phase}
\label{subsec:disaster_phase}

\begin{figure}[htp]
    \centering
    \begin{subfigure}[t]{0.32\linewidth}
        \centering
        \includegraphics[width=\linewidth,height=4cm]{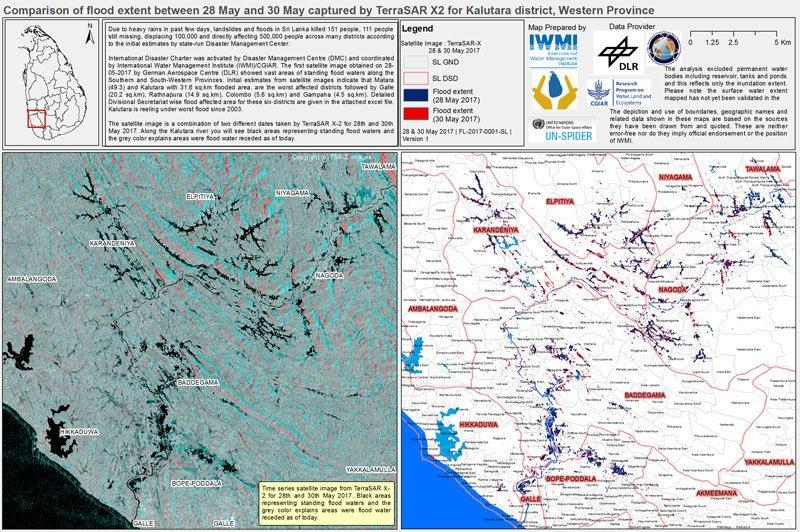}
        \caption{The \#flood extent in Kalutara District, \#SriLanka, was captured by \#TerraSARX on 30 May}
    \end{subfigure}
    \hfill
    \begin{subfigure}[t]{0.32\linewidth}
        \centering
        \includegraphics[width=\linewidth,height=4cm]{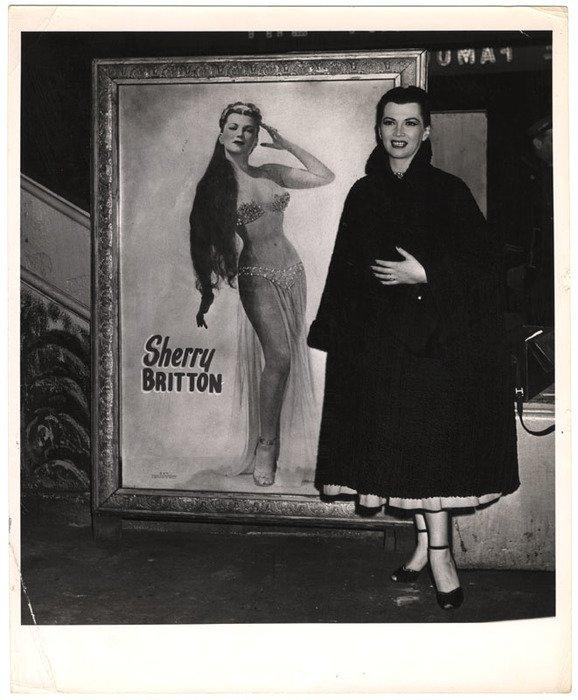}
        \caption{She got her stage name from a liquor store on a bottle of Harvey's Bristol Cream Sherry}
    \end{subfigure}
    \hfill
    \begin{subfigure}[t]{0.32\linewidth}
        \centering
        \includegraphics[width=\linewidth,height=4cm]{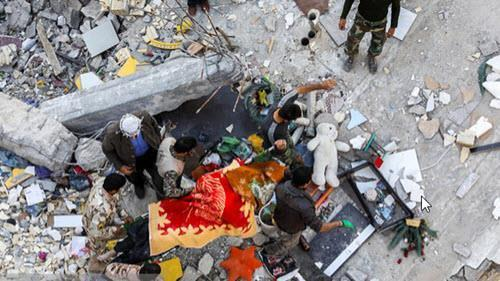}
        \caption{\#foxnews \#cnn \#ap \#Iran National Television (INTV) donates to the victims of the earthquake}
    \end{subfigure}

    \begin{subfigure}[t]{0.32\linewidth}
        \centering
        \includegraphics[width=\linewidth,height=4cm]{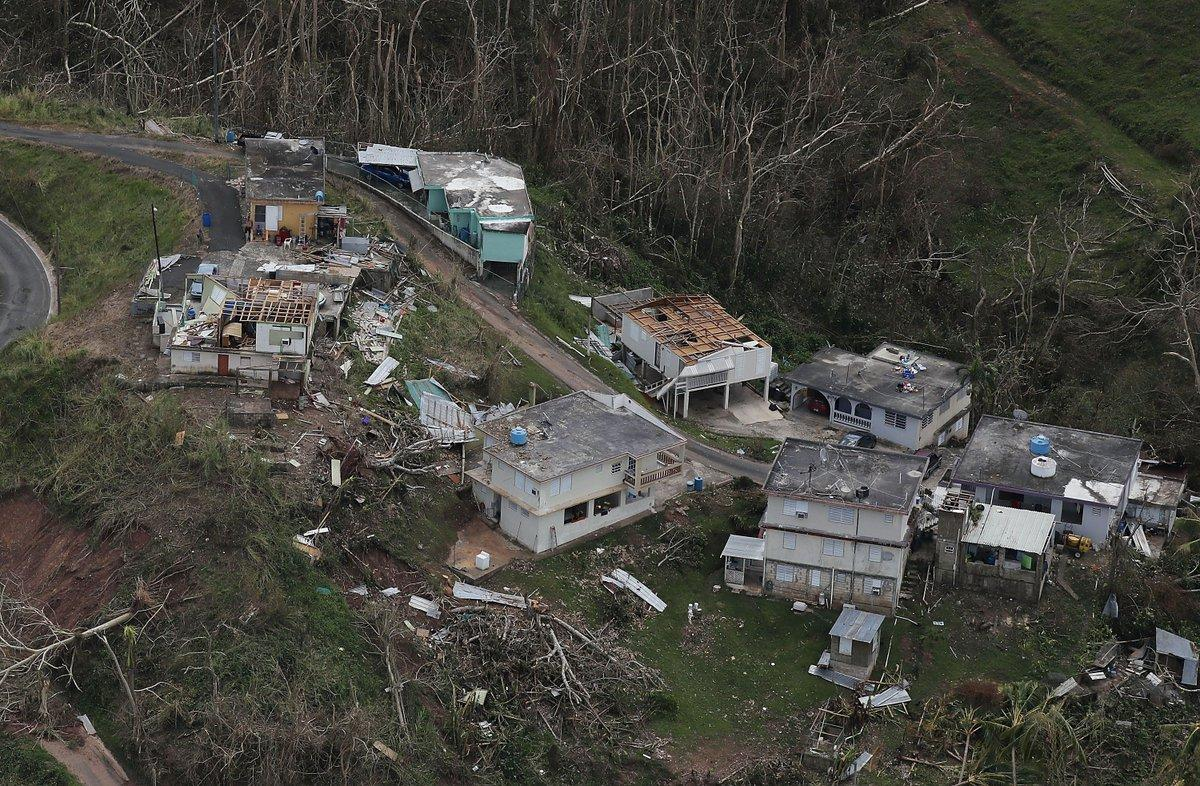}
        \caption{J.J. Barea teamed up with Mark Cuban to bring aid to Puerto Rico}
    \end{subfigure}
    \hfill
    \begin{subfigure}[t]{0.32\linewidth}
        \centering
        \includegraphics[width=\linewidth,height=4cm]{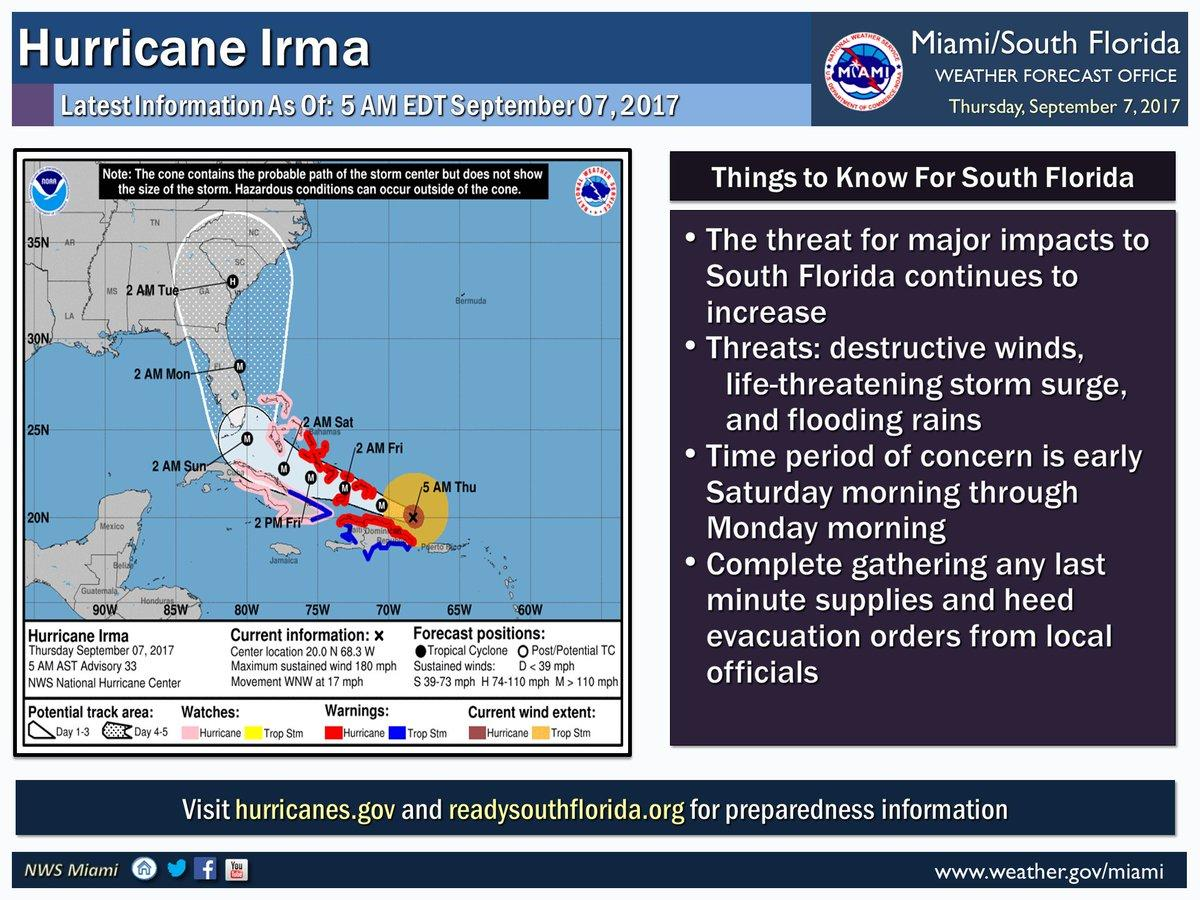}
        \caption{5AM Thursday 9/7 Hurricane Irma update: The threat for major impacts for South Florida continues to increase. \#FLwx}
    \end{subfigure}
    \hfill
    \begin{subfigure}[t]{0.32\linewidth}
        \centering
        \includegraphics[width=\linewidth,height=4cm]{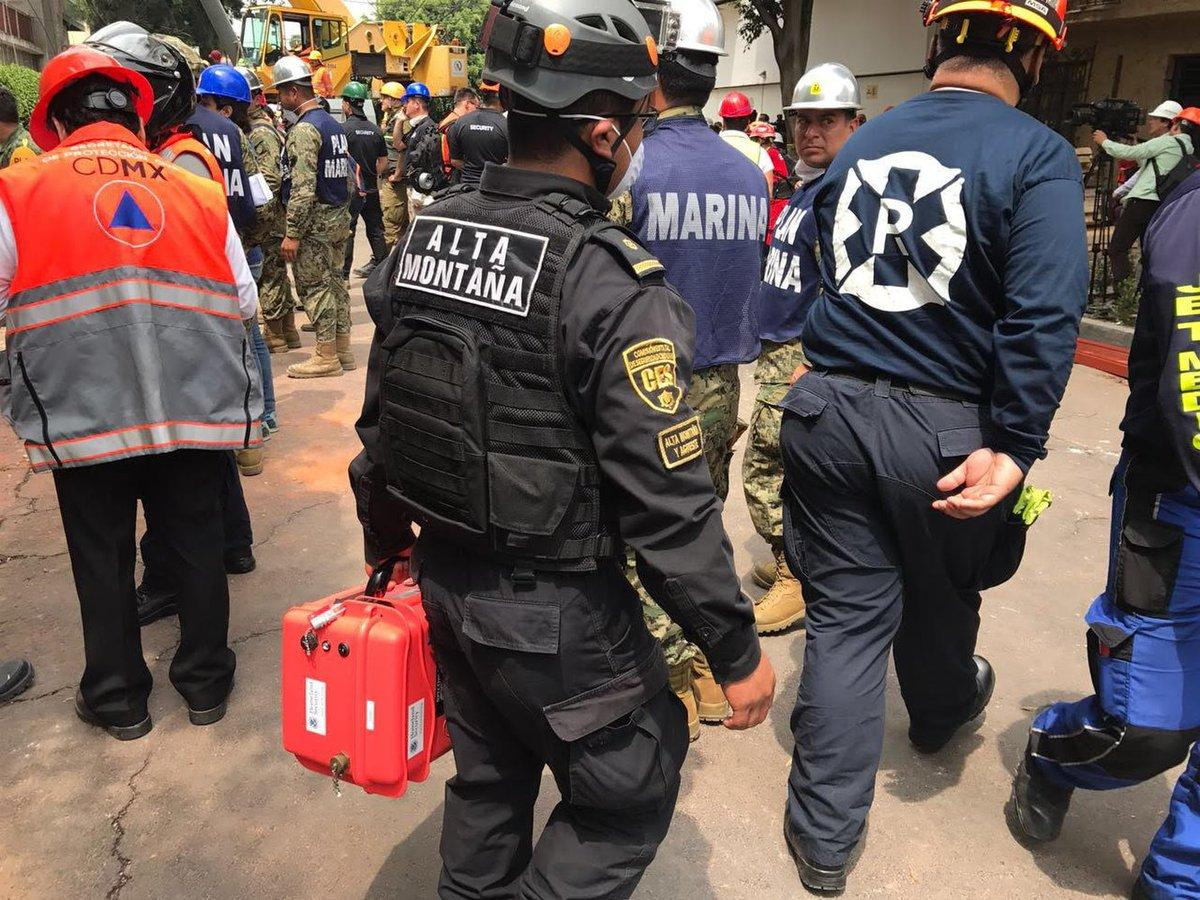}
        \caption{NASA Tech Is Helping Find Mexico Earthquake Victims Buried in Rubble}
    \end{subfigure}


    \caption{Example data from the CrisisMMD dataset, used to train Enabler agents across Scenario Levels 1, 2, and 3 corresponding to the following data categories: (a) Informative, (b) Not-informative, (c) Affected individuals, (d) Infrastructure/utility damage, (e) Other relevant information, and (f) Rescue/volunteering efforts, 
    }
    \label{fig:crisismmd_data}
\end{figure}

From the proposed framework in Figure \ref{fig:systemDiagram}, the decisions from \verb|Level|-1, \verb|Level|-2, and \verb|Level|-3 are for activities in the disaster phase related to emergency response services. 

The CrisisMMD \cite{FirojAlam2018} dataset comprises image+text pairs with their corresponding annotations, as seen in Figure \ref{fig:crisismmd_data}. This dataset is utilized to train the three \verb|Enabler| agents for the three \verb|Levels|. The \verb|Enabler| agent here is a multimodal multiclass classification model.

\begin{itemize}
    \item For \verb|Level|-1, the \verb|Enabler| agent is trained to classify the image-text pair as either `informative' or `not-informative' regarding the disaster.
    
    \item For \verb|Level|-2, the \verb|Enabler| agent identifies the type of humanitarian aid represented by the image-text pair. The target classes are `affected individuals,' `infrastructure/utility damage,' `other relevant information,' and `rescue/volunteering efforts.' The labels `injured or dead people' and `missing or found people' were relabeled to `affected individuals,' and `vehicle damage' was relabeled as `infrastructure/utility damage.' The `not humanitarian' label was removed due to its irrelevance in the decision process.
    
    \item For \verb|Level|-3, the \verb|Enabler| agent assesses the severity of damage in the image-text pair. The target classes are `little or no damage', and `severe damage'. The label `mild damage' was relabeled to `little or no damage' to better reflect the context.
\end{itemize}

\begin{figure}[htp]
    \centering
    \includegraphics[width=1.0\linewidth]{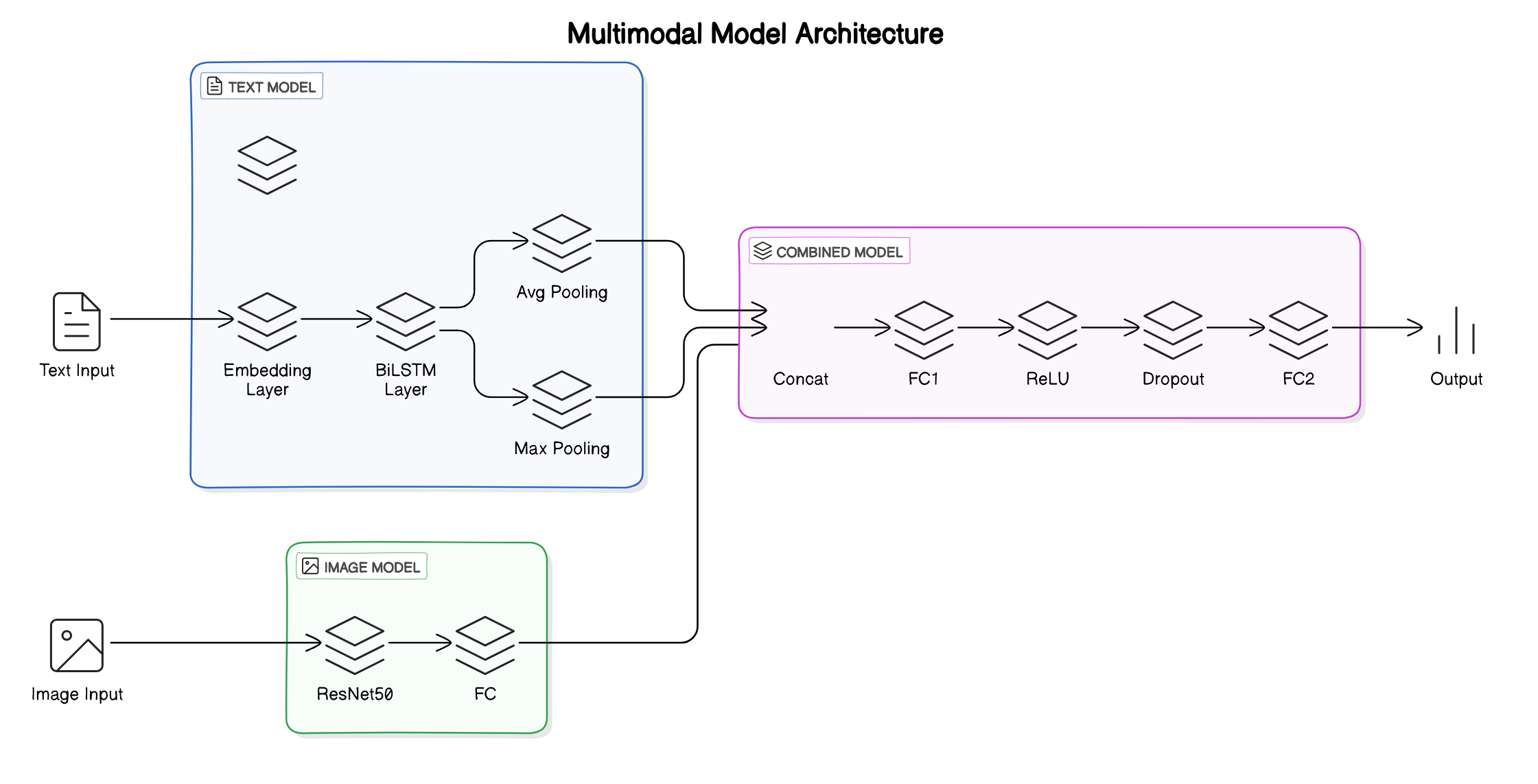}
    \caption{Multimodal model architecture}
    \label{fig:multimodalArch}
\end{figure}

Inspired by \cite{Koshy2023}, a multimodal model architecture consisting of a BiLSTM for text processing and a ResNet50 for image processing is designed, as seen in Figure \ref{fig:multimodalArch}. The BiLSTM produces two sets of hidden states, one from processing the text forwards \(\overrightarrow{h_t}\) and another from processing it backwards \(\overleftarrow{h_t}\). These hidden states are concatenated to form the final hidden state for each time step \(h_t = [\overrightarrow{h_t}, \overleftarrow{h_t}]\). The sequence of hidden states is then summarized using average pooling,
\[
\text{avg\_pool} = \frac{1}{T} \sum_{t=1}^{T} h_t,
\]
and max pooling,
\[
\text{max\_pool} = \max_{t=1}^{T} h_t,
\]
with the resulting vectors concatenated to form the text feature vector.  
Example: suppose \(T{=}3\) and each \(h_t\in\mathbb{R}^3\) is  
\(h_1=[0.2,\,0.7,\,-0.1],\;h_2=[0.4,\,0.1,\,0.5],\;h_3=[-0.3,\,0.6,\,0.0]\).  
Element-wise averaging gives  
\(\text{avg\_pool}=\frac{h_1+h_2+h_3}{3}=[(0.2+0.4-0.3)/3,\,(0.7+0.1+0.6)/3,\,
(-0.1+0.5+0.0)/3]=[0.1,\,0.467,\,0.133]\).  
Element-wise max pooling yields  
\(\text{max\_pool}=[0.4,\,0.7,\,0.5]\).  
Concatenating them produces a \(2\times3=6\)-dimensional text vector \([0.1,\,0.467,\,0.133,\,0.4,\,0.7,\,0.5]\). 
This vector is then concatenated with the 256-dimensional image feature vector obtained from the ResNet50 model \cite{ResNet2015}, forming a combined feature vector. The combined vector, with a dimensionality of \(2 \times \text{hidden\_size} + 256\), is passed through a fully connected layer with ReLU activation and dropout to reduce overfitting. Finally, the processed features are fed into another fully connected layer, producing the final output; the confidence scores for all classes are recorded as an array and stored as the judgement used later in a \verb|Scenario|.

In preparing the dataset for training the \verb|Enabler| agents, the Twitter text is cleaned to replace contractions with their full forms, standardize numbers, remove special characters, clean up social media-specific elements (like URLs, mentions, and hashtags), and remove common stop-words, this allows for a more uniform and simplified text.

A Keras Tokenizer \cite{tensorflow_tokenizer} is fit on the cleaned Twitter text data, converting the text into sequences of integers that correspond to the words in the tokenizer's vocabulary. The tokenized sequences are then padded to have the same length which is necessary for consistent input to machine learning models. The Keras tokenizer uses a vocabulary size of 20,000 for \verb|Level|-1 and \verb|Level|-2, and 10,000 for \verb|Level|-3. In all cases, padded with a maximum sequence length of 20.

Additionally, pre-trained GloVe embeddings \cite{glove2014}  are loaded that map each word in the tokenizer's vocabulary to its corresponding GloVe vector, enabling the model to leverage these rich word representations during training. The GloVe embeddings \cite{pennington-etal-2014-glove} help capture semantic relationships between words. They are pre-trained on a large corpus of text capturing a broad understanding of word relationships and meanings. They help in reducing the complexity of the data reducing the risk of overfitting, and they can help mitigate the problem of out-of-vocabulary (OOV) words.

The datasets for each \verb|Level| in the disaster phase are stratified into a train set and validation set in an 80:20 ratio such that the proportion of each class in the target variable is preserved in both sets, as seen in Table \ref{tab:level1_cls_distrb}. 

The images are resized to 224x224, and random horizontal flip augmentation is applied using the TorchVision \cite{torchvision2016} library. Normalization is then performed to adjust the pixel values so that each colour channel has a mean of 0 and a standard deviation of 1.

The learning rates are 0.0001 and the number of epochs is 100 for the three \verb|Levels|. The batch sizes are 4 for \verb|Level|-1, and 32 for \verb|Level|-2 and \verb|Level|-3.

\begin{table}[htp]
\caption{Class distribution for disaster phase labels in Level-1, Level-2, and Level-3.}
\label{tab:level1_cls_distrb}
\centering
\begin{tabular}{lcccc}
\hline
\textbf{Label} & \textbf{Total Data Count} \\
\hline
\textbf{Level-1} & \multicolumn{1}{c}{} \\
\hline
Informative & 9162 \\
Not Informative & 8817 \\
\hline
\textbf{Level-2} & \multicolumn{1}{c}{-} \\
\hline
Infrastructure and Utility Damage & 3830 \\
Rescue Volunteering or Donation Effort & 2231 \\
Other Relevant Information & 2528 \\
Affected Individuals & 686 \\
\hline
\textbf{Level-3} & \multicolumn{1}{c}{} \\
\hline
Severe Damage & 2212 \\
Little or No Damage & 1314 \\
\hline
\end{tabular}
\end{table}

\subsubsection{Post-Disaster Phase}
\label{subsec:postdisaster-phase}

The decisions of damage assessment made at \verb|Level|-4 and \verb|Level|-5 are crucial components of the post-disaster phase. In this phase, the \verb|Enabler| is provided with images captured by satellites and drones. The \verb|Enabler|, which is a multilabel image classification model designed to evaluate the severity of the damage depicted in the images. The xBD and RescueNet datasets, which are utilized for this task, provide segmentation annotations for each corresponding image. 

\begin{itemize}
    \item For \verb|Level|-4, the \verb|Enabler| agent is trained to identify `no damage' and `major damage' from the satellite images from the xBD dataset. The labels `destroyed' and `minor damage' were relabelled as `major damage' to simplify training, and avoid confusion among labels of damage.
    \item  For \verb|Level|-5, the \verb|Enabler| agent is trained to identify `building destroyed' and `building no damage' from the drone images from RescueNet dataset. The label `building major damage' was relabelled to `building destroyed', and the `building minor damage' was relabelled to `building no damage'. The following classes: `background', `water', `vehicle', `tree', `pool', `road clear', and `road blocked' were ignored by replacing pixel values with 0 using their segmentation annotations, this was done to simplify the decisions to be made at this \verb|Level|.
\end{itemize}

Although training a segmentation model would potentially yield higher accuracy, a multilabel classification approach is chosen, to simplify the training process. This is justified by the fact that an image can be associated with multiple labels. The model used for multilabel classification is built upon the ResNet-50 architecture, which uses deep convolutional layers and residual connections. Using a pre-trained ResNet-50 model as the base, takes advantage of the knowledge it has acquired from large datasets like ImageNet to extract meaningful features from the input images. To accommodate the multilabel classification task, the original fully connected layer of the ResNet-50 was replaced with a new fully connected layer that outputs a number of features corresponding to the total number of classes in the dataset. This modification allows the model to predict multiple labels for a single image. Additionally, a sigmoid activation function was applied to the output of this modified layer. The sigmoid function converts the raw outputs into probabilities, enabling the independent prediction of each class label, which is essential for multilabel classification tasks. This customized ResNet-50 model, integrated with a sigmoid activation function, serves as the \verb|Enabler| agent and is trained on a well-structured dataset that accurately reflects the complex and multifaceted nature of post-disaster damage assessment.

\begin{figure}[htp]
    \centering
    \begin{subfigure}[t]{0.32\linewidth}
        \centering
        \includegraphics[width=\linewidth,height=4cm]{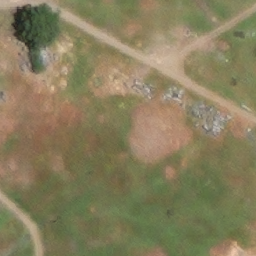}
        \caption{No Damage}
    \end{subfigure}
    \hfill
    \begin{subfigure}[t]{0.32\linewidth}
        \centering
        \includegraphics[width=\linewidth,height=4cm]{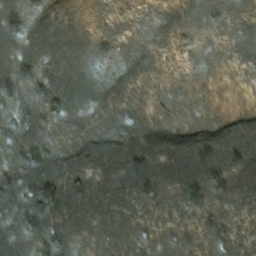}
        \caption{No Damage}
    \end{subfigure}
    \hfill
    \begin{subfigure}[t]{0.32\linewidth}
        \centering
        \includegraphics[width=\linewidth,height=4cm]{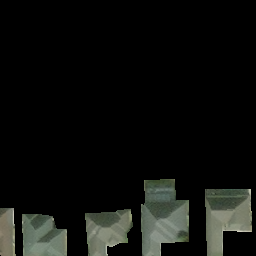}
        \caption{Major Damage}
    \end{subfigure}

    \begin{subfigure}[t]{0.32\linewidth}
        \centering
        \includegraphics[width=\linewidth,height=4cm]{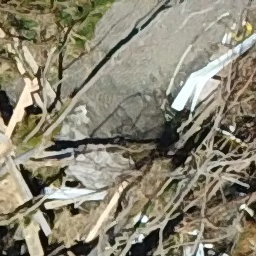}
        \caption{Building No Damage}
    \end{subfigure}
    \hfill
    \begin{subfigure}[t]{0.32\linewidth}
        \centering
        \includegraphics[width=\linewidth,height=4cm]{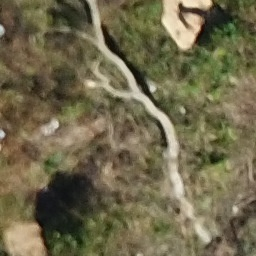}
        \caption{Building No Damage}
    \end{subfigure}
    \hfill
    \begin{subfigure}[t]{0.32\linewidth}
        \centering
        \includegraphics[width=\linewidth,height=4cm]{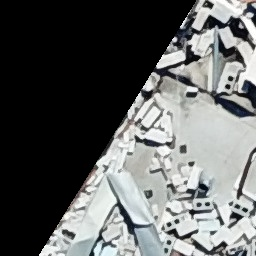}
        \caption{Building Destroyed}
    \end{subfigure}

    \caption{Example data from the post-disaster phase, displaying the images with their corresponding annotation, used to train Enabler agents across Scenario Levels 3 and 4. (a), (b), and (c) 
    are satellite images from the xBD dataset, (d), (e), and (f) 
    are drone-captured images from the RescueNet dataset.}
    \label{fig:postdisaster_data}
\end{figure}

Given that the dataset contains images with large resolutions, it is essential to ensure efficient training and faster learning. As part of this study, the images are divided into smaller patches of size 256x256 with a step size of 256, as seen in Figure \ref{fig:postdisaster_data}. To construct the dataset for training the multilabel classification model, the segmentation annotations were used to classify the images and generate a CSV file for both training and validation. The CSV file contains columns representing each class, as well as the image paths. Each image is recorded only once in the CSV file, and if an image is associated with multiple labels, the corresponding columns for those labels are populated with a value of 1, indicating the presence of those labels in the image. This method effectively structures the dataset for training the multilabel classification model. The train and validation dataset is constructed from the train and validation data provided from the original xBD dataset and RescueNet dataset, this is seen in Table \ref{tab:level2_cls_distrb}.

The learning rates are 0.001 for \verb|Level|-4, and 0.0001 for \verb|Level|-5. The number of epochs is 100 and the batch sizes are 16 for the two \verb|Levels|.

\begin{table}[htp]
\caption{Class distribution for post-disaster phase labels in Level-4, and Level-5.}
\label{tab:level2_cls_distrb}
\centering
\begin{tabular}{lcccc}
\hline
\textbf{Label} & \textbf{Train Data Count} & \textbf{Validation Data Count} \\
\hline
\textbf{Level-4} & \multicolumn{1}{c}{} \\
\hline
No Damage & 89566 & 29856 \\
Major Damage & 38601 & 12355 \\
\hline
\textbf{Level-5} & \multicolumn{1}{c}{} \\
\hline
Building No-Damage & 46569 & 1974 \\
Building Destroyed & 27951 & 1232 \\
\hline
\end{tabular}
\end{table}

\subsubsection{Enabler Agent Metrics}
\label{sec:metrics_classification}

To evaluate the performance of the Enabler Agent model for the classification task, the following metrics were computed using the `scikit-learn' \cite{scikit-learn} package in Python:
\begin{itemize}
    \item {\bf Precision}: The proportion of true positive predictions out of all positive predictions.
    \item {\bf Recall}: The proportion of true positive predictions out of all actual positive instances in the dataset.
    \item {\bf F1-Score}: The harmonic mean of precision and recall.
    \item {\bf Accuracy}: The overall proportion of correct predictions out of all predictions made by the model.
    \item {\bf Macro Avg}: The average of a metric (e.g., precision, recall, F1-Score) calculated independently for each class and then averaged.
    \item {\bf Weighted Avg}: The average of a metric (e.g., precision, recall, F1-Score) calculated independently for each class, weighted by the number of true instances for each class.
    \item {\bf Support}: The number of true instances for each class in the dataset.
\end{itemize}
The primary metric used to select the best model is the {\bf Macro Average F1-score}, as it provides the model's performance across the classes by averaging the F1-scores for each class, and is well suited due to the imbalance in the dataset.

\subsection{Reinforcement Learning Decision Maker}
The \verb|Decision Maker| agent determines the activities to be undertaken at each \verb|Level| in a \verb|Scenario|. This section discusses the reinforcement learning \verb|Decision Maker| agent, and the following section will discuss the human operator agent.

The reinforcement learning (RL) agent is what brings about autonomous decision-making in the framework. The judgements made by the \verb|Enabler| agents at each \verb|Level| inform the decision-making of the RL agent, which brings structure to the decision-making process, making it reliable and justifiable. This decision flow of the RL agent can be seen in Figure \ref{fig:systemDecision}.

The data for training and evaluation of the RL agent is prepared by running inference using the trained \verb|Enabler| agents for each \verb|Level| in a \verb|Scenario| on their respective training and validation datasets, and then storing the predicted confidence scores for all classes along with the ground truth. These results are then utilized by the RL agent. The RL agent's environment is a custom Gymnasium \cite{OpenAIGym2016}, and is designed so that, when a decision is required at a \verb|Level|, an unused random data record is fetched from the stored outputs for the \verb|Enabler| agent at that \verb|Level|. Since each \verb|Level| has a varying number of possible actions that are expected, it is required that the RL agent environment has a different action space defined for each \verb|Level|. One episode in the RL environment corresponds to a \verb|Scenario|, with each step representing a \verb|Level| within that \verb|Scenario| based on the action taken.

The observation space for the RL agent is a one-dimensional array with a length of 9, a breakdown of the observation space is seen in Table \ref{tab:observation_space}. The first 4 elements consist of the prediction confidence array from the \verb|Enabler| agent for that \verb|Level|. Since the prediction confidence array can vary in length across different \verb|Levels|, it is padded with zeros if its length is less than 4. The remaining 5 elements form the \verb|Level| vector, which indicates the specific \verb|Level| from which the observation is collected. For example, `10000' represents \verb|Level|-1, and `00001' represents \verb|Level|-5. The final observation space is the concatenation of the padded prediction confidence array and the \verb|Level| vector.

\begin{table}[htp!]
\caption{Observation space}
\label{tab:observation_space}
\centering
\begin{tabular}{|l|c|}
\hline
\textbf{Component}         & \textbf{Dimension} \\ \hline
Expected Length of Prediction Confidence Array      & 4                  \\ \hline
Length of \verb|Level| Vector Array & 5                  \\ \hline
\textbf{Total Observation} & \textbf{9}         \\ \hline
\end{tabular}
\end{table}

Each \verb|Level| offers a varying number of possible actions, as shown in Table \ref{tab:action_space}. Correct or incorrect actions can result in advancing to the next \verb|Level|, with associated rewards or consequences. The RL environment is structured so that the action space is determined by the current \verb|Level|. At the end of each step, if the current \verb|Level| is completed, the action space for the next \verb|Level| is set, ensuring that invalid actions from other \verb|Levels| are not predicted by the RL agent. Additionally, each \verb|Level| includes a \verb|Gather Additional Data| action, which allows the \verb|Decision Maker| agent to request new data for better decision-making while remaining at the current \verb|Level|. This action does not advance the agent to the next \verb|Level| but instead updates the observation space with new data, keeping the agent on the same \verb|Level| for the next step. 

\begin{table}[htp!]
\caption{Action spaces and corresponding actions for each level}
\label{tab:action_space}
\centering
\begin{tabular}{|l|c|c|}
\hline
\textbf{Task} & \textbf{Action Space (Discrete)} & \textbf{Actions} \\ \hline
\verb|Level-1| & 3 & 0, 1, 2 \\ \hline
\verb|Level-2| & 5 & 0, 1, 2, 3, 4 \\ \hline
\verb|Level-3| & 3 & 0, 1, 2 \\ \hline
\verb|Level-4| & 3 & 0, 1, 2 \\ \hline
\verb|Level-5| & 3 & 0, 1, 2 \\ \hline
\end{tabular}
\end{table}

\begin{itemize}
    \item For \verb|Level|-1, the action space includes:  0 which represents `informative', 1 which represents `not informative', and 2 which represents \verb|Gather Additional Data|.
    \item For \verb|Level|-2, the action space includes:  0 which represents`affected individuals', 1 which represents `infrastructure and utility damage', 2 which represents `other relevant information', 3 which represents `rescue and volunteering efforts', and 4 which represents \verb|Gather Additional Data|.
    \item For \verb|Level|-3, the action space includes:  0 which represents `little or no damage', 1 which represents `severe damage', and 2 which represents \verb|Gather Additional Data|.
    \item For \verb|Level|-4, the action space includes:  0 which represents `no damage', 1 which represents `major damage', and 2 which represents \verb|Gather Additional Data|.
    \item For \verb|Level|-5, the action space includes:  0 which represents `building no damage', 1 which represents `building destroyed', and 2 which represents \verb|Gather Additional| \verb|Data|.
\end{itemize}

In a \verb|Scenario|, each \verb|Level| has an allocation of 5 credits to request additional data, requesting additional data can result in a minor penalty of \verb|-1|. Correct decisions result in a reward of \verb|+1| and wrong decisions result in a penalty of \verb|-5|. 

An Advantage Actor--Critic (A2C) algorithm~\cite{A2C2016} with a multilayer-perceptron (MLP) policy was trained using the following hyper-parameters.  
The discount factor was fixed to $\gamma = 0.995$, emphasising delayed rewards.  
Each policy update processed $n_{\text{steps}} = 128$ environment interactions, a compromise between sample efficiency and return-variance reduction.  
The entropy coefficient was set to \textit{ent\_coef} = 0.02 to foster exploration, while the value-function loss weight was kept at $v\!f_{\text{coef}} = 0.5$ to balance actor and critic objectives.  
Gradients were clipped at \textit{max\_grad\_norm} = 1 to avoid destabilising updates.  
Optimisation employed Adam with a learning rate of $5\times10^{-4}$ and $\varepsilon = 10^{-7}$.  
Training ran for $8.0\times10^{7}$ time-steps, with evaluation and logging performed every 1\,000 steps. The hyper-parameters of the A2C agent were tuned in two sequential stages.

\begin{enumerate}[label=\arabic*.]
\item \textbf{Initialisation from established practice.}  
      Default settings recommended by Stable-Baselines3 for long-horizon, discrete-action problems were adopted as the starting point:  
      $\gamma=0.99,\;n_{\text{steps}}=128,\;\text{ent\_coef}=0.01,\;vf_{\text{coef}}=0.5,\;\text{max\_grad\_norm}=0.5,\;\text{lr}=3\times10^{-4}$.
\item \textbf{Coarse grid search.}  
      A grid search comprising 20 distinct hyper-parameter combinations (five random seeds each) was run for 5 M time-steps per combination.  
      The following ranges were explored while all other knobs were kept at the Stage-1 values:  
      \[
        \begin{aligned}
        \gamma            &\in \{0.95,\, 0.99,\, 0.995,\, 0.999\},\\
        \textit{ent\_coef} &\in \{0,\, 0.01,\, 0.02,\, 0.05\},\\
        \textit{lr}        &\in \{1\times10^{-3},\, 5\times10^{-4},\, 1\times10^{-4}\}.
        \end{aligned}
      \]
      Performance was measured by mean episodic reward on a held-out validation set of 500 scenarios; the configuration with the highest score was selected for full training.
\end{enumerate}

The search identified the following setting as optimal:  
\(\gamma=0.995\) (indicating a strong preference for delayed rewards),  
\(n_{\text{steps}}=128\) (balancing Generalised Advantage Estimation accuracy against memory use),  
\(\textit{ent\_coef}=0.02\) (yielding an $\approx 3\%$ macro-$F_{1}$ gain over 0.01 and 0.05 by promoting exploration),
\(vf_{\text{coef}}=0.5\) (higher values slowed policy learning with no accuracy benefits),  
\(\text{max\_grad\_norm}=1\) (clipping at 0.5 caused under-fitting; unclipped runs diverged), and  
\(\text{lr}=5\times10^{-4}\) with Adam (\(\varepsilon=10^{-7}\)).  

The fixed hyper-parameter set was subsequently trained for the full 80M time-steps, with evaluation and logging performed every 1000 steps. The RL decision process at each \verb|Level| was modelled as a finite Markov Decision Process $\langle\mathcal S,\mathcal A,T,R,\gamma\rangle$:

\begin{itemize}
\item \textbf{State} $s_t\!=\![c_1,c_2,c_3,c_4,\;l_1,\dots,l_5,\;q_t]\in\mathcal S\subset\mathbb R^{10}$  
      combines three elements  
      \begin{enumerate}[label=(\roman*)]
      \item the \emph{prediction–confidence vector} $\mathbf c_t$ (padded to length~4);  
      \item the one-hot \emph{level indicator} $\mathbf l_t$ (length~5);  
      \item the scalar \emph{credit counter} $q_t\in\{0,\dots,5\}$ that records how many “gather–data” credits remain at the current level.  
      \end{enumerate}
      A concrete example for Level-2 might be  
      $s_t=[0.71,0.22,0.05,0.00,\;0,1,0,0,0,\;4]$,  
      meaning the \verb|Enabler| is moderately confident the post concerns \emph{infrastructure/utility damage} (class-1), we are at Level-2, and 4 credits are still available.

\item \textbf{Action} $a_t\in\mathcal A(\mathbf l_t)$ comes from the discrete sets in Table \ref{tab:action_space}; the action space therefore changes deterministically with the current level.

\item \textbf{Transition} $T(s_{t+1}\!\mid s_t,a_t)$ is deterministic:  
      \begin{itemize}
      \item if $a_t$ is a classification label and \textit{correct}, advance to the next level and reset $q_{t+1}=5$;  
      \item if $a_t$ is \textit{incorrect}, the episode terminates;  
      \item if $a_t$ is \verb|Gather Additional Data|, stay at the same level, decrement $q_{t+1}=q_t-1$, and refresh $\mathbf c_{t+1}$ with a new sample.
      \end{itemize}

\item \textbf{Reward} is a scalar  
\[
R(s_t,a_t)=
\begin{cases}
+1, & \text{correct label}\\[4pt]
-5, & \text{incorrect label}\\[4pt]
-1, & \mathtt{Gather\ Additional\ Data}
\end{cases}
\]  
      Together with $\gamma=0.995$, the agent is encouraged to maximise long-term accuracy while sparingly using its limited credits.

\item \textbf{Episode termination} occurs when the agent either (i) finishes Level-5, (ii) makes an incorrect decision, or (iii) exhausts all five credits without making a final classification.
\end{itemize}

Assume the agent is at Level-3 with $q_t=2$ credits left and receives  
$\mathbf c_t=[0.10,0.85,0.05,0.00]$.  
If it predicts \emph{severe damage} ($a_t=1$) and that is indeed the ground truth, it obtains $R=+1$ and transitions to Level-4 with $q_{t+1}=5$.  
If instead it chooses \verb|Gather Additional Data| ($a_t=2$), it pays $R=-1$, stays at Level-3 with $q_{t+1}=1$, and a fresh image–text pair is sampled.

\subsection{Human Operator Decision Maker}

This section discusses the human operator \verb|Decision Maker| agent, which determines the activities to be undertaken at each \verb|Level| in a \verb|Scenario|. The human operators are the participants who were recruited for the evaluation study. A web application called \verb|Disaster Maestro| (\url{https://disaster-maestro.web.app/}) 
was developed to collect human responses and evaluate their performance in comparison to the previous reinforcement learning agent. Unlike the RL agent, human operators were not provided with judgment insights from the \verb|Enabler| agents, as seen in Figure \ref{fig:systemDecision}. Instead, they had to independently assess and make decisions, simulating the decision-making process of a stakeholder in disaster management. This design aimed to replicate real-world conditions where stakeholders must often rely on their own judgment without the aid of advanced analytical tools.


To ensure a robust evaluation of the RL agent, responses were gathered from a diverse group of participants, as per the following inclusion criteria: 
victims affected by disaster, volunteers involved in rescue or relief operations, or stakeholders in disaster management.
Participants were recruited via networking platforms such as Facebook and LinkedIn, as well as through close contacts. The recruitment process involved searching hashtags related to disasters on social media and reaching out to individuals who had posted relevant content. The individuals were victims/volunteers/stakeholders in the 2018 floods in Kerala, typhoon Gaemi 
in Taiwan, and 2024 Hualien earthquake. 
This diverse input brought significant depth and realism to the evaluation process. The participants were provided with the URL to the web application, and asked to complete at least 2 \verb|Scenarios| before they chose to end the survey. 
\begin{itemize}
    \item In \verb|Level|-1, \verb|Level|-2, and \verb|Level|-3 participants are presented with image-text pairs from the preprocessed CrisisMMD dataset and are asked to make a judgment based on the data. They then indicate their decision by clicking the button that corresponds to their decision.
    \item In \verb|Level|-4 participants are presented with images captured by satellites from the preprocessed xBD dataset and are asked to make a judgment based on the data. They then indicate their decision by clicking the button that corresponds to their decision.
    \item In \verb|Level|-5 participants are presented with images captured by drones from the preprocessed RescueNet dataset and are asked to make a judgment based on the data. They then indicate their decision by clicking the button that corresponds to their decision.
\end{itemize}

\begin{figure}[htp]
    \begin{subfigure}[b]{0.49\linewidth} 
        \centering
        \includegraphics[width=\linewidth]{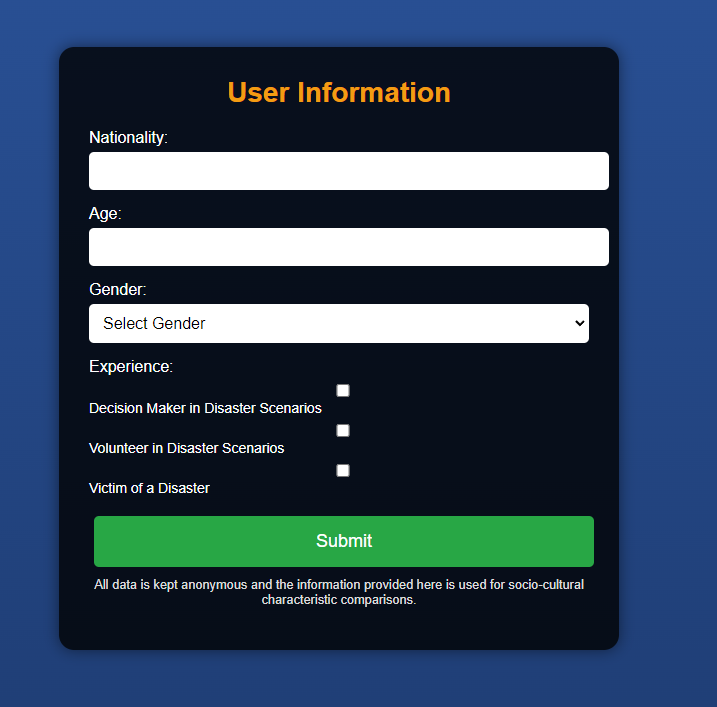}
        \caption{GUI displaying the user form for information collection.}
    \end{subfigure}
    \hfill
    \begin{subfigure}[b]{0.49\linewidth} 
        \centering
        \includegraphics[width=\linewidth]{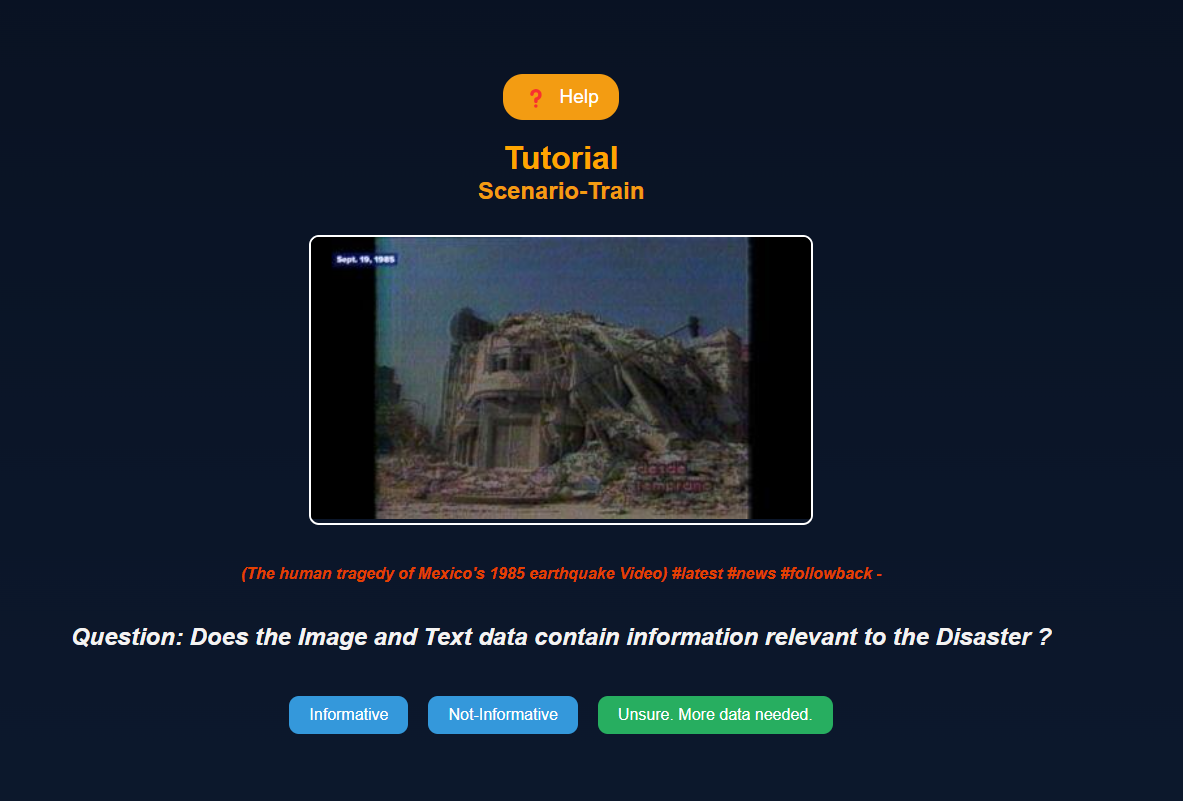}
        \caption{GUI showing tutorial screen.}
    \end{subfigure}

    \begin{subfigure}[t]{0.49\linewidth} 
        \centering
        \includegraphics[width=\linewidth]{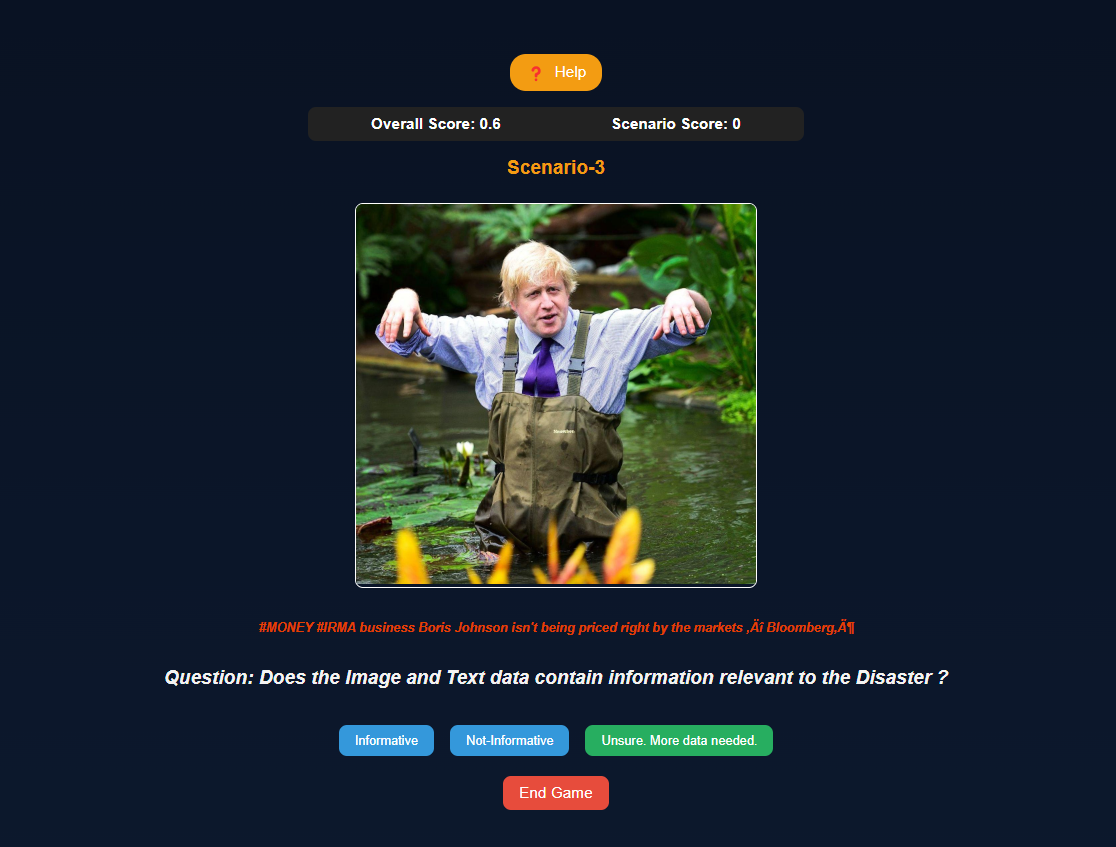}
        \caption{Example GUI screen with scoring.}
    \end{subfigure}
    \hfill
    \begin{subfigure}[t]{0.49\linewidth} 
        \centering
        \includegraphics[width=\linewidth]{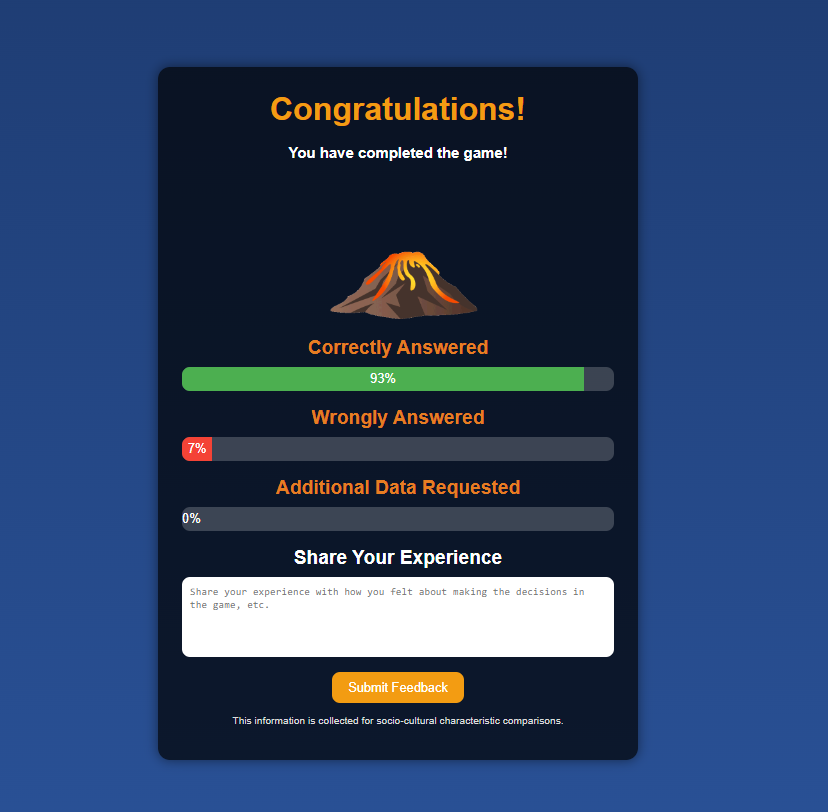}
        \caption{Example GUI of results screen.}
    \end{subfigure}
    
    \caption{Example GUI screens of the web application}
    \label{fig:gui_web}
\end{figure}

The data prepared for training and evaluating the RL agent is uploaded to a cloud storage to be utilized for the web application. The web application was carefully structured to guide users, as seen in Figure \ref{fig:gui_web}.

The process begins with an introduction to the project, providing context and explaining the purpose of the exercise. Following this, a training section allows users to interact with the system and learn how to make informed decisions before they are scored. This section is crucial in helping participants understand the complexities of disaster management and the types of decisions they would need to make. In the application logic, users are presented with a series of \(N\) \verb|Scenarios|, the \verb|Scenarios| shown in the application come from the validation set, so that there is consistency when evaluating the human responses with the RL agent. The application computes the participants' final score and showcases insights on the correctness of their decisions. No personal data was collected from the participants of this study. In the web application, the participants are shown: 
\begin{itemize}
    \item For \verb|Level|-1, data which can be `informative' or `not informative', where they inherently judge the data and then select an action `informative', `not informative', or \verb|Gather Additional Data|.
    \item For \verb|Level|-2, data which can represent `affected individuals', `infrastructure and utility damage', `other relevant information', or `rescue and volunteering efforts', where they judge the data and then select an action: `affected individuals', `infrastructure and utility damage', `other relevant information', `rescue and volunteering efforts', or \verb|Gather Additional Data|.
    \item For \verb|Level|-3, data which can indicate `little or no damage' or `severe damage', where they judge the extent of damage and then select an action: `little or no damage', `severe damage', or \verb|Gather Additional Data|.
    \item For \verb|Level|-4, data which can indicate `no damage' or `major damage', where they judge the extent of damage and then select an action: `no damage', `major damage', or \verb|Gather Additional Data|.
    \item For \verb|Level|-5, data which can indicate `building no damage' or `building destroyed', where they judge the extent of damage and then select an action: `building no damage', `building destroyed', or \verb|Gather Additional Data|.
\end{itemize}
The web application is built in ReactJS and hosted on Firebase, making use of the database and cloud storage to store the responses from the participants, and to store the data records to display the questions to the user in the survey. 
In a \verb|Scenario|, each \verb|Level| has an allocation of 5 credits to request additional data, requesting additional data can result in a minor penalty of \verb|-1|. Correct decisions result in a reward of \verb|+1| and wrong decisions result in a penalty of \verb|-5|. 

\subsection{Decision Maker metrics}
\label{sec:rl_metrics}
To evaluate the performance of the \verb|Decision Maker| agents in both cases, reinforcement learning or human operator, the following metrics were calculated:
\begin{itemize}
    \item \verb|tree_score|: The tree score is the total reward accumulated in one episode (one \verb|Scenario|). The score has a range of [-5,+5], where a score of +5 indicates that the \verb|Decision Maker| agent made correct decisions at all \verb|Levels| within that \verb|Scenario|.  This metric quantifies how well the \verb|Decision Maker| agent performs across different \verb|Scenario|.
    \item \verb|isTreeCorrectlyAnswered|: This metric is calculated as the mean number of correct decisions made at each \verb|Level| in a \verb|Scenario|, providing a confidence score of the \verb|Decision Maker| agent's performance. A value of 1.0 indicates that the agent made correct decisions at all \verb|Levels| in a \verb|Scenario|. 
    \item \verb|isTreeWronglyAnswered|: This metric is calculated by the mean number of wrong decisions made at each \verb|Level| in a \verb|Scenario|, providing a confidence score of the \verb|Decision Maker| agent's performance. A value of 1.0 indicates that the agent made wrong decisions at all \verb|Levels| in a \verb|Scenario|. 
    \item \verb|isGatherAdditionalDataRequested|: This metric is calculated by the mean number of times the gather additional data action was utilized at each \verb|Level| in a \verb|Scenario|, providing an analysis of the \verb|Decision Maker| agent's performance. A value of 1.0 indicates that the agent requested additional data at all \verb|Levels| in a \verb|Scenario|.
\end{itemize}
For all the metrics, the mean and standard deviation are calculated at regular intervals, with an interval of 1000 steps.

\section{Results}\label{sec5}
This section presents our study's experimental results. First, a brief overview of the \verb|Enabler| agents' results is provided, followed by the core of the results which is the comparison of the performance of the autonomous \verb|Decision Maker| agent to that of a human operator.

\subsection{Enabler Agent}
This section discusses the experimental results for the \verb|Enabler| agents for each \verb|Level| in a \verb|Scenario|, using the metrics introduced in Section \ref{sec:metrics_classification}.

\FloatBarrier                 
\needspace{14\baselineskip} 
\subsubsection{Level-1}
\begin{figure}[htbp]
    \centering
    \includegraphics[width=0.6\linewidth,height=0.5\linewidth]{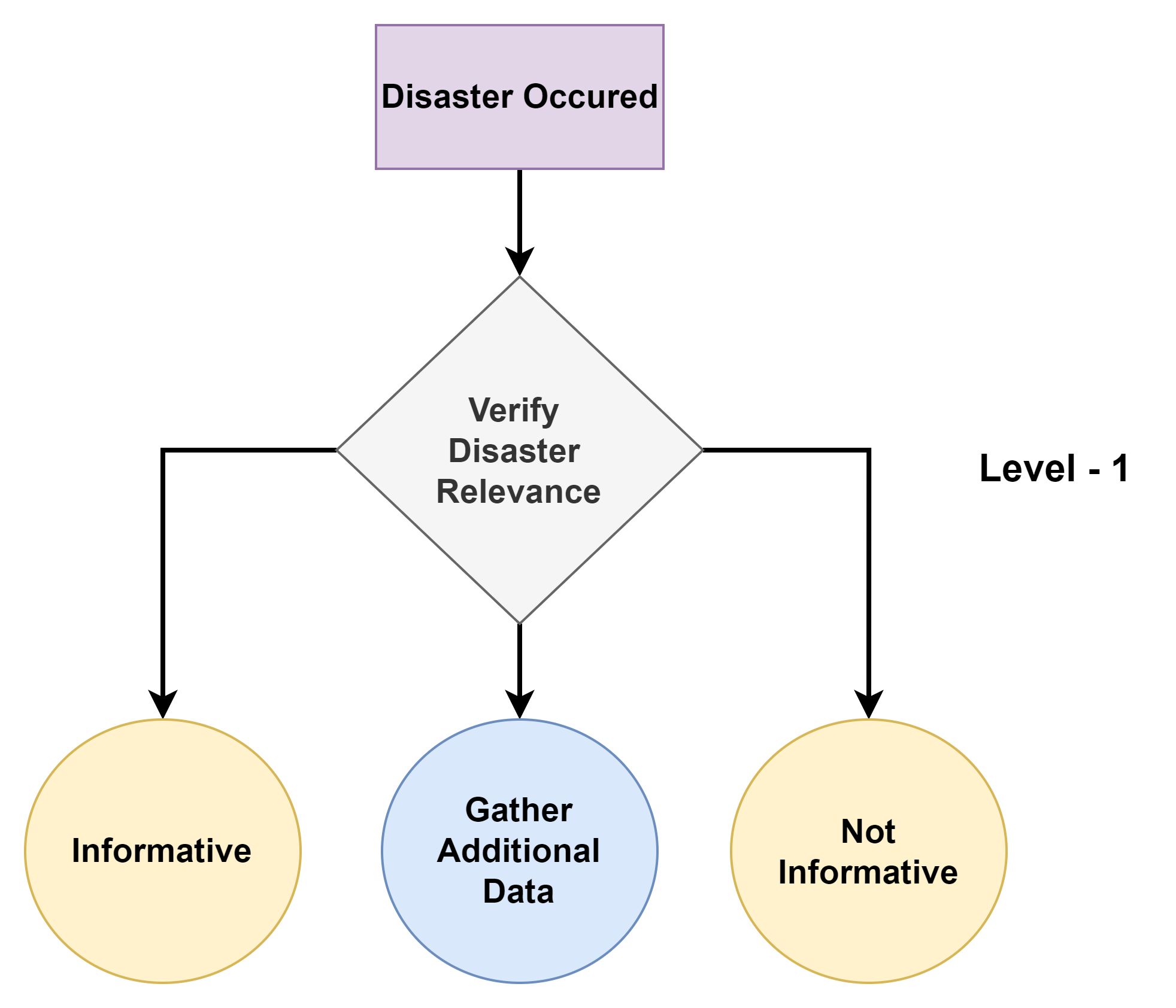}
    \caption{Overview of Level-1.}
    \label{fig:level1-overview}
\end{figure}
\begin{table}[htp]
\caption{Performance metrics for the level-1 enabler agent.}
\label{tab:level1_metrics}
\centering
\begin{tabular}{lcccc}
\hline
\textbf{Label} & \textbf{Precision} & \textbf{Recall} & \textbf{F1-Score} & \textbf{Support} \\
\hline
Informative (0) & 0.8665 & 0.7730 & 0.8171 & 1855 \\
Not Informative (1) & 0.7832 & 0.8731 & 0.8257 & 1742 \\
\hline
\textbf{Accuracy} & \multicolumn{1}{c}{0.8215} \\
\hline
\textbf{Macro Avg} & 0.8248 & 0.8231 & 0.8214 & 3597 \\
\textbf{Weighted Avg} & 0.8261 & 0.8215 & 0.8213 & 3597 \\
\hline
\end{tabular}
\end{table}

In \verb|Level|-1, the \verb|Enabler| agent judges the image + text data to verify its relevance in relation to the disaster, as seen in Figure \ref{fig:level1-overview}, classifying the results into 2 classes: `informative' and `not informative'. The performance metrics for the best model are shown in Table \ref{tab:level1_metrics}.
From the table, a Macro Average F1-score of 82.14\% indicates that the model's performance is balanced across both classes.

\FloatBarrier                 
\needspace{14\baselineskip} 
\subsubsection{Level-2}
\begin{figure}[htbp]
    \centering
    \includegraphics[width=\linewidth,height=0.5\linewidth]{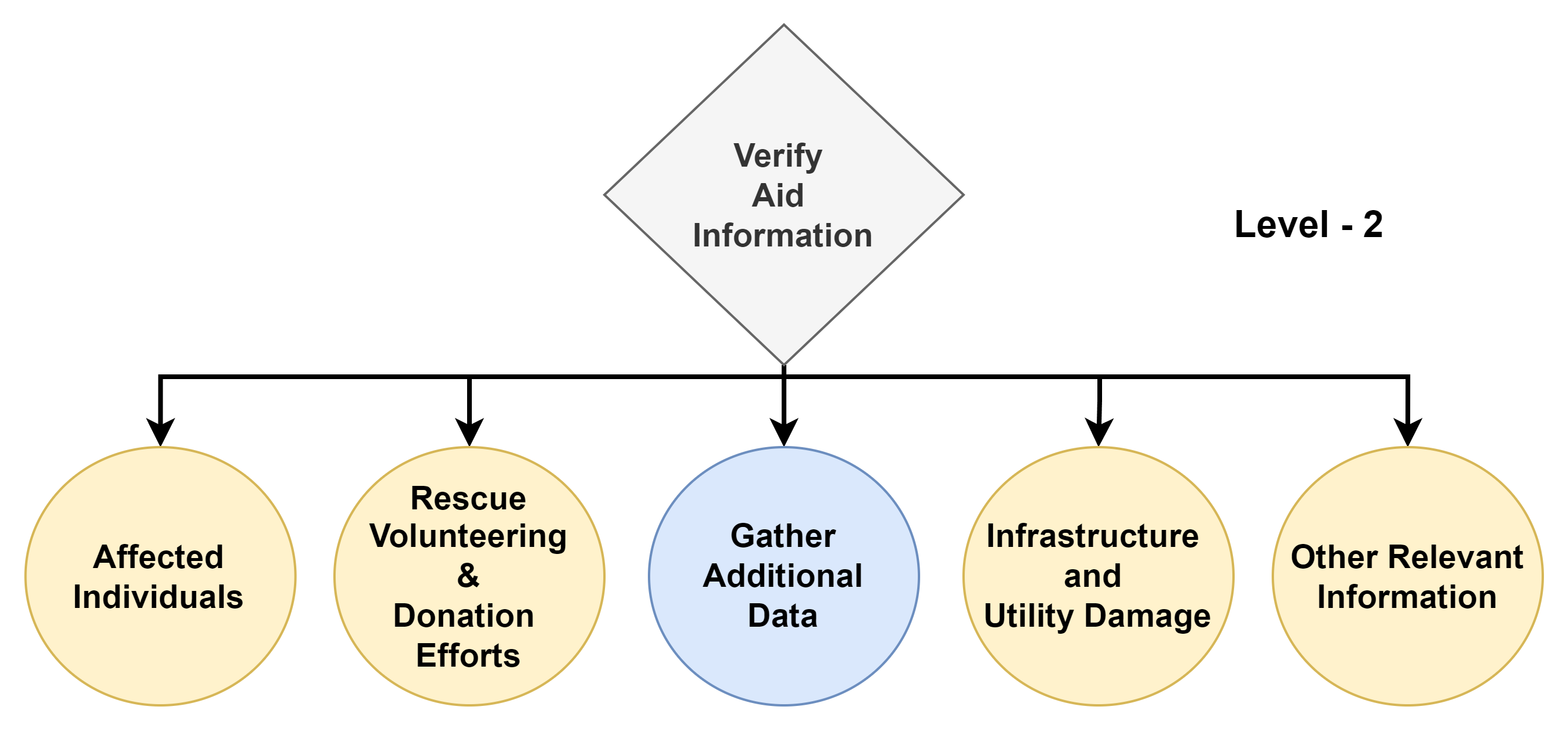}
    \caption{Overview of Level-2.}
    \label{fig:level2-overview}
\end{figure}
\begin{table}[htp]
\caption{Performance metrics for the Level-2 Enabler Agent.}
\label{tab:level2_metrics}
\centering
\begin{tabular}{lcccc}
\hline
\textbf{Label} & \textbf{Precision} & \textbf{Recall} & \textbf{F1-Score} & \textbf{Support} \\
\hline
Affected Individuals (0) & 0.4667 & 0.5109 & 0.4878 & 137 \\
Infrastructure and Utility Damage (1) & 0.8486 & 0.8930 & 0.8702 & 766 \\
Other Relevant Information (2) & 0.9158 & 0.8814 & 0.8983 & 506 \\
Rescue/Volunteering/Donation Effort (3) & 0.7476 & 0.6906 & 0.7179 & 446 \\
\hline
\textbf{Accuracy} & \multicolumn{1}{c}{0.8129} \\
\hline
\textbf{Macro Avg} & 0.7447 & 0.7440 & 0.7436 & 1855 \\
\textbf{Weighted Avg} & 0.8145 & 0.8129 & 0.8130 & 1855 \\
\hline
\end{tabular}
\end{table}

In \verb|Level|-2, the \verb|Enabler| agent evaluates image-text data to verify its relevance to disaster aid, as illustrated in Figure \ref{fig:level2-overview}. The data is classified into four categories: `affected individuals', `infrastructure and utility damage', `other relevant information', and `rescue, volunteering, or donation efforts'. The performance metrics for the best model are presented in Table \ref{tab:level2_metrics}.
The model achieves a moderate Macro Average F1-score of 74.36\%, indicating reasonably balanced performance across the four classes. Notably, the model performs best on `other relevant information' with an F1-score of 89.83\%, and `infrastructure and utility damage' with an F1-score of 87.02\%, while the performance on `affected individuals' is lower, with an F1-score of 48.78\%, but this is expected due to lower of number of samples available to train for the `affected individuals' class. 
\FloatBarrier                 
\needspace{14\baselineskip} 
\subsubsection{Level-3}
\begin{figure}[htbp]
    \centering
    \includegraphics[width=0.7\linewidth,height=0.5\linewidth]{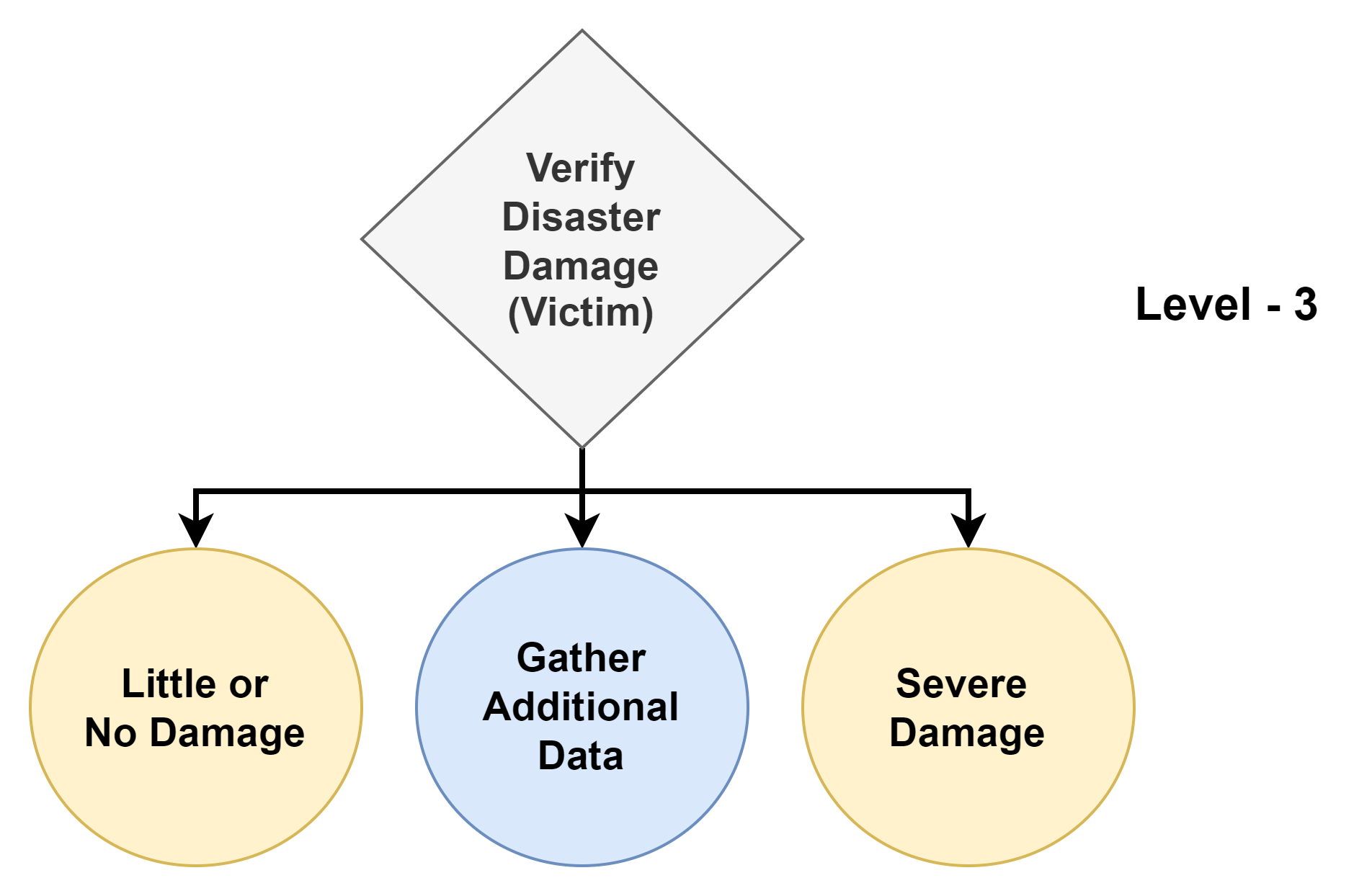}
    \caption{Overview of Level-3.}
    \label{fig:level3-overview}
\end{figure}
\begin{table}[htp]
\caption{Performance metrics for the Level-3 Enabler Agent}
\label{tab:level3_metrics}
\centering
\begin{tabular}{lcccc}
\hline
\textbf{Label} & \textbf{Precision} & \textbf{Recall} & \textbf{F1-Score} & \textbf{Support} \\
\hline
Little or No Damage (0) & 0.6438 & 0.7148 & 0.6775 & 263 \\
Severe Damage (1) & 0.8188 & 0.7652 & 0.7911 & 443 \\
\hline
\textbf{Accuracy} & \multicolumn{1}{c}{0.7465} \\
\hline
\textbf{Macro Avg} & 0.7313 & 0.7400 & 0.7343 & 706 \\
\textbf{Weighted Avg} & 0.7536 & 0.7465 & 0.7488 & 706 \\
\hline
\end{tabular}
\end{table}

In \verb|Level|-3, the \verb|Enabler| agent judges the image + text data to verify its disaster damage captured by victims in relation to the disaster, as seen in Figure \ref{fig:level3-overview}. The model is trained for 2 classes: `little or no damage' and `severe damage'. The performance metrics for the best model are shown in Table \ref{tab:level3_metrics}.
From the table, a Macro Average F1-score of 73.43\% indicates reasonably balanced performance across both classes. 


\FloatBarrier                 
\needspace{14\baselineskip} 
\subsubsection{Level-4}
\begin{figure}[htbp]
    \centering
    \includegraphics[width=0.7\linewidth,height=0.5\linewidth]{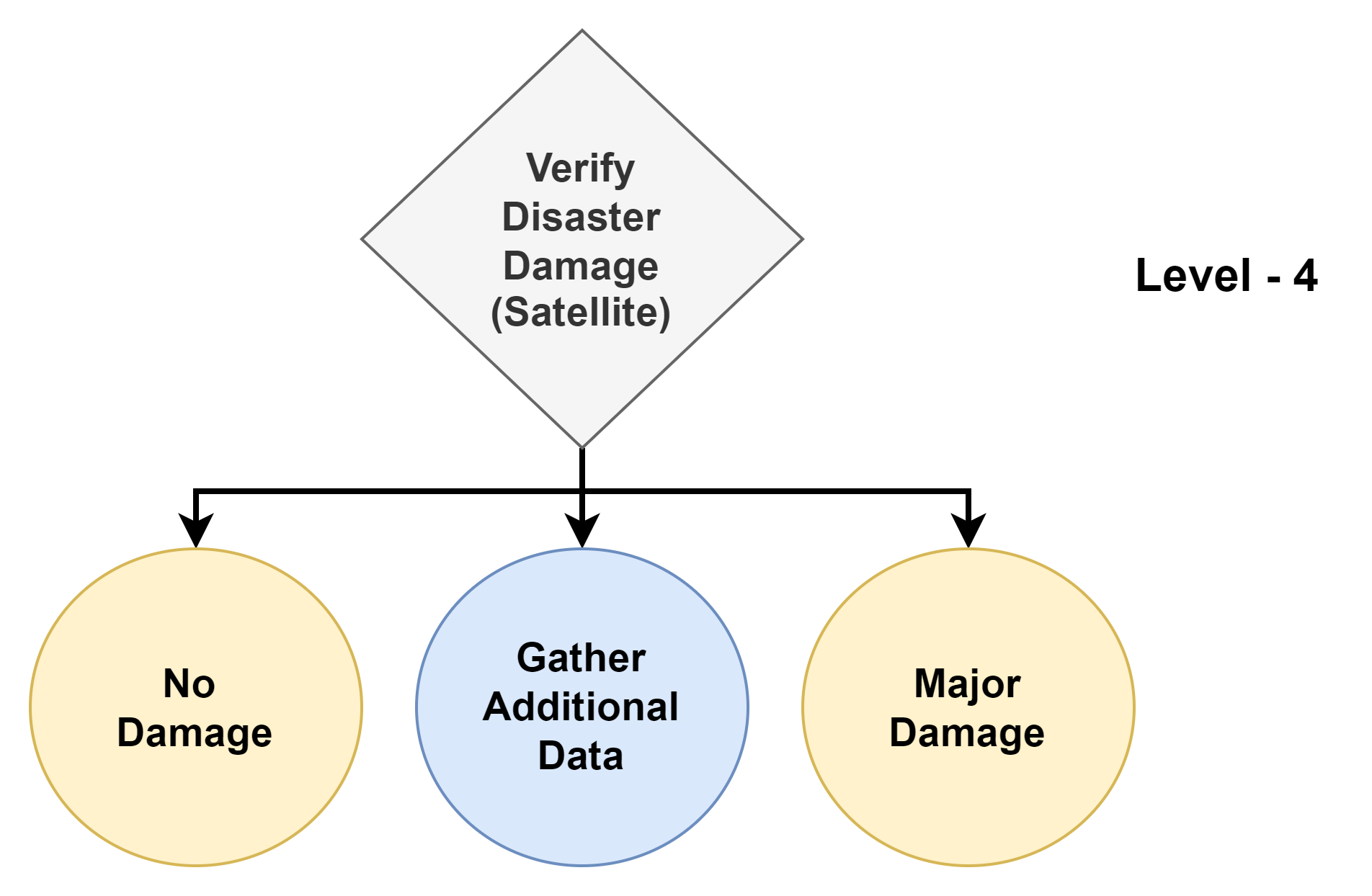}
    \caption{Overview of Level-4.}
    \label{fig:level4-overview}
\end{figure}
\begin{table}[htp]
\caption{Performance metrics for Level-4 Enabler Agent}
\label{tab:level4_metrics}
\centering
\begin{tabular}{lcccc}
\hline
\textbf{Label} & \textbf{Precision} & \textbf{Recall} & \textbf{F1-Score} & \textbf{Support} \\
\hline
No Damage & 1.0000 & 1.0000 & 1.0000 & 29856 \\
Major Damage & 0.9985 & 0.9938 & 0.9961 & 12355 \\
\hline
\textbf{Micro Avg} & 0.9996 & 0.9982 & 0.9989 & 42211 \\
\textbf{Macro Avg} & 0.9993 & 0.9969 & 0.9981 & 42211 \\
\textbf{Weighted Avg} & 0.9996 & 0.9982 & 0.9989 & 42211 \\
\hline
\end{tabular}
\end{table}

In \verb|Level|-4, the \verb|Enabler| agent judges the image to verify its disaster damage captured by satellites in relation to the disaster, as seen in Figure \ref{fig:level4-overview}. The classification model is trained for 2 classes: `no damage' and `major damage'. The performance metrics for the best model are shown in Table \ref{tab:level4_metrics}. From the table, a Macro Average F1-score of 99.89\%  indicates the model's performance is balanced across both classes. The model correctly identified 99.38\% of `major damage' instances with a precision of 99.85\%. Similarly, the model achieved 100\% recall for `no damage' instances with a precision of 100\%. 

\FloatBarrier                 
\needspace{14\baselineskip} 
\subsubsection{Level-5}
\begin{figure}[htbp]
    \centering
    \includegraphics[width=0.7\linewidth,height=0.5\linewidth]{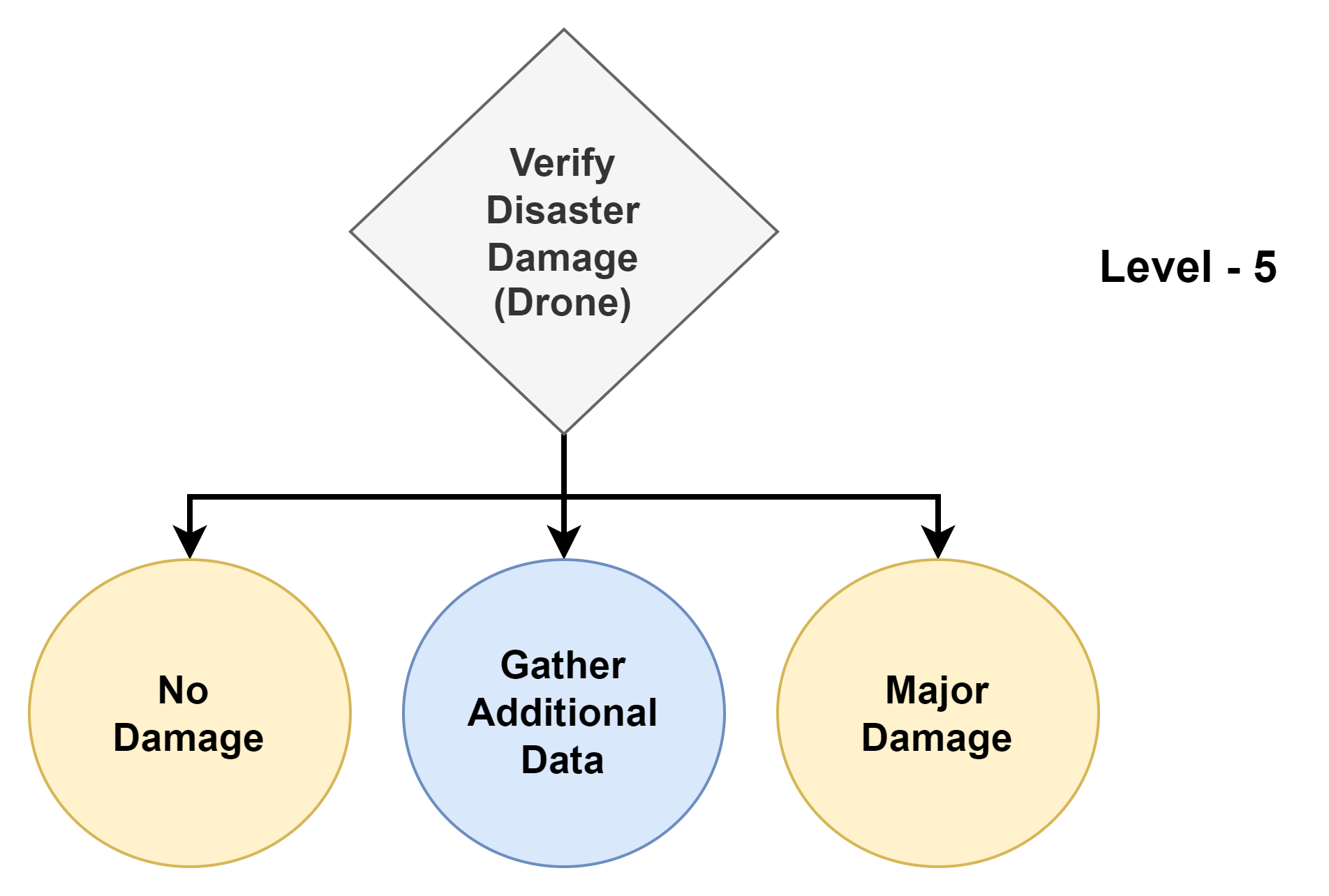}
    \caption{Overview of Level-5.}
    \label{fig:level5-overview}
\end{figure}
\begin{table}[htp]
\caption{Performance metrics for Level-5 Enabler Agent}
\label{tab:level5_metrics}
\centering
\begin{tabular}{lcccc}
\hline
\textbf{Label} & \textbf{Precision} & \textbf{Recall} & \textbf{F1-Score} & \textbf{Support} \\
\hline
Building No Damage & 0.7475 & 0.7244 & 0.7358 & 1974 \\
Building Destroyed & 0.7176 & 0.4724 & 0.5698 & 1232 \\
\hline
\textbf{Micro Avg} & 0.7386 & 0.6276 & 0.6786 & 3206 \\
\textbf{Macro Avg} & 0.7326 & 0.5984 & 0.6528 & 3206 \\
\textbf{Weighted Avg} & 0.7360 & 0.6276 & 0.6720 & 3206 \\
\hline
\end{tabular}
\end{table}

In \verb|Level|-5, the \verb|Enabler| agent judges the image to verify its disaster damage captured by drones in relation to the disaster, as seen in Figure \ref{fig:level5-overview}. The classification model is trained for 2 classes: `building no damage' and `building destroyed'. The performance metrics for the best model are shown in Table \ref{tab:level5_metrics}. From the table, a Macro Average F1-score of 65.28\% indicates reasonably balanced performance across both classes. The model correctly identified 72.44\% of `building no damage' instances with a precision of 71.76\%. Similarly, the model achieved 47.24\% recall for `building destroyed' instances with a precision of 71.76\%, the model finds it hard to find all instances of building destroyed, which can be attributed to the similar features between the classes, resulting in confusion among the classes. 

\subsubsection{Results Summary for Enabler Agents}

\begin{table}[htp]
\caption{Results summary for the enabler agents trained for each level in a scenario.}
\label{tab:results_summary_enabler}
\centering
\begin{tabular}{lcc}
\hline
\makecell{\textbf{Enabler Agent}} & \makecell{\textbf{Macro Avg. F1-Score}} \\
\hline
\textbf{Level-1} & 0.8214 \\
\textbf{Level-2} & 0.7436 \\
\textbf{Level-3} & 0.7343 \\
\textbf{Level-4} & 0.9989 \\
\textbf{Level-5} & 0.6528 \\
\hline
\end{tabular}
\end{table}

The performance of the \verb|Enabler| agents across different levels in the structured decision-making framework for disaster management varies, as reflected by the Macro Average F1-Scores in Table \ref{tab:results_summary_enabler}. \verb|Level|-1, responsible for classifying data as informative or not, achieved a balanced performance with an F1-Score of 82.14\%. \verb|Level|-2, which identifies relevant humanitarian efforts, showed moderate performance with an F1-Score of 74.36\%. \verb|Level|-3, focused on assessing damage severity based on victim-captured data, attained a reasonably balanced F1-Score of 73.43\%. The highest performance was observed in \verb|Level|-4, where satellite imagery was used to distinguish between major and no damage, achieving an exceptional F1-Score of 99.89\%. In contrast, \verb|Level|-5, which deals with drone-captured data, exhibited the lowest performance with an F1-Score of 65.28\%, indicating challenges in distinguishing between undamaged and destroyed buildings. Since the primary objective of the framework is to demonstrate autonomous decision-making rather than the performance of the \verb|Enabler| agents, the performance is satisfactory. It effectively provides judgment insights on the data to inform the RL \verb|Decision Maker| agent. 

\subsection{Decision Maker Agent}
This section presents the experimental results from training and evaluating the \verb|Decision| \verb|Maker| agents across multiple \verb|Scenarios|, using the metrics introduced in Section \ref{sec:rl_metrics}. The metric plots are generated with a 10\% running average smoothing to highlight learning trends.

\subsubsection{Reinforcement Learning}

\begin{figure}[htp]
    \begin{subfigure}[t]{0.49\linewidth} 
        \centering
        \includegraphics[width=\linewidth]{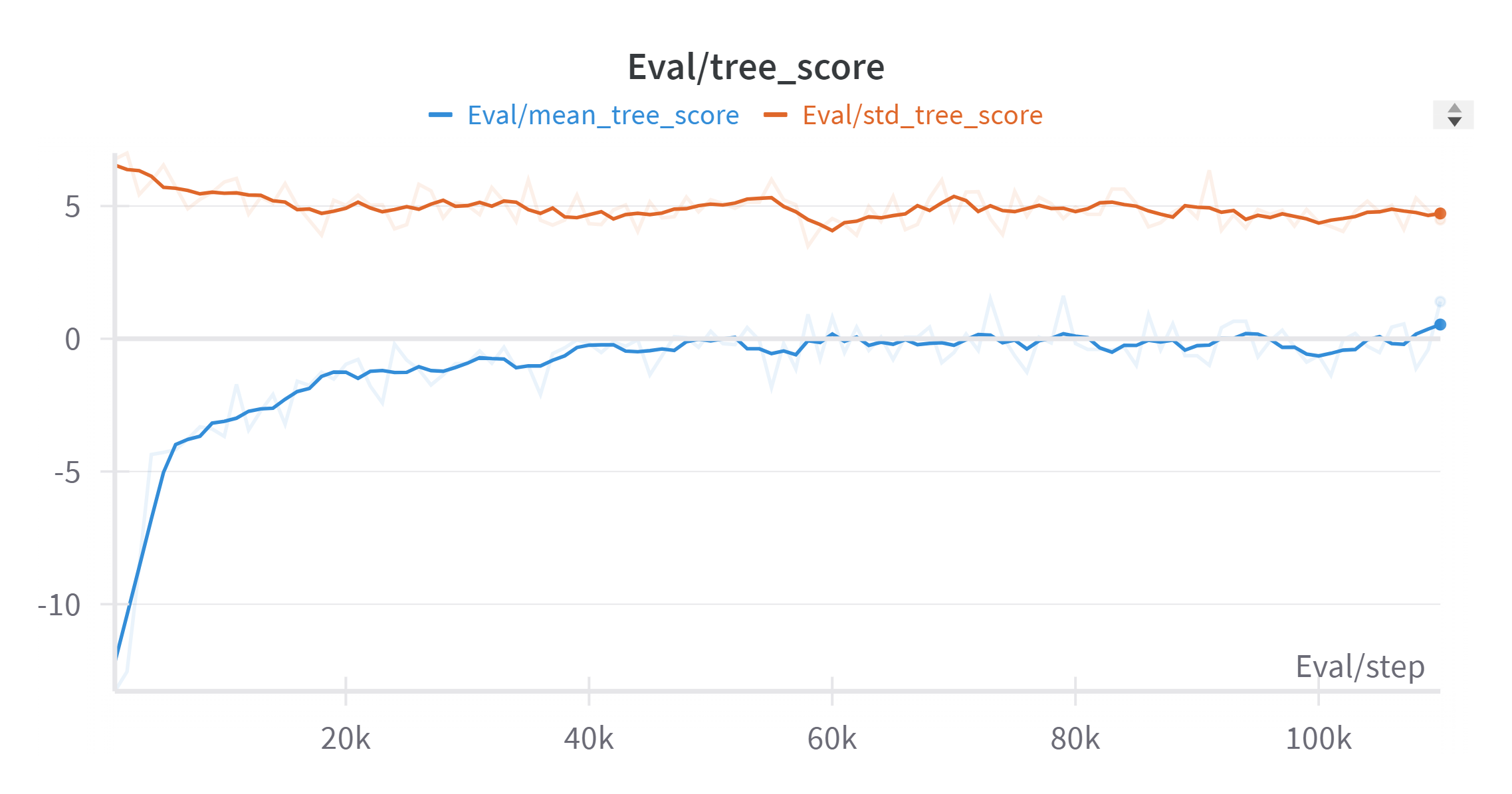}
        \caption{\small (Eval): \(\mu\)=1.4 and \(\sigma\)=4.49 of `tree\_score'}
        \label{subfig:rl-fig1}
    \end{subfigure}
    \hfill
    \begin{subfigure}[t]{0.49\linewidth} 
        \centering
        \includegraphics[width=\linewidth]{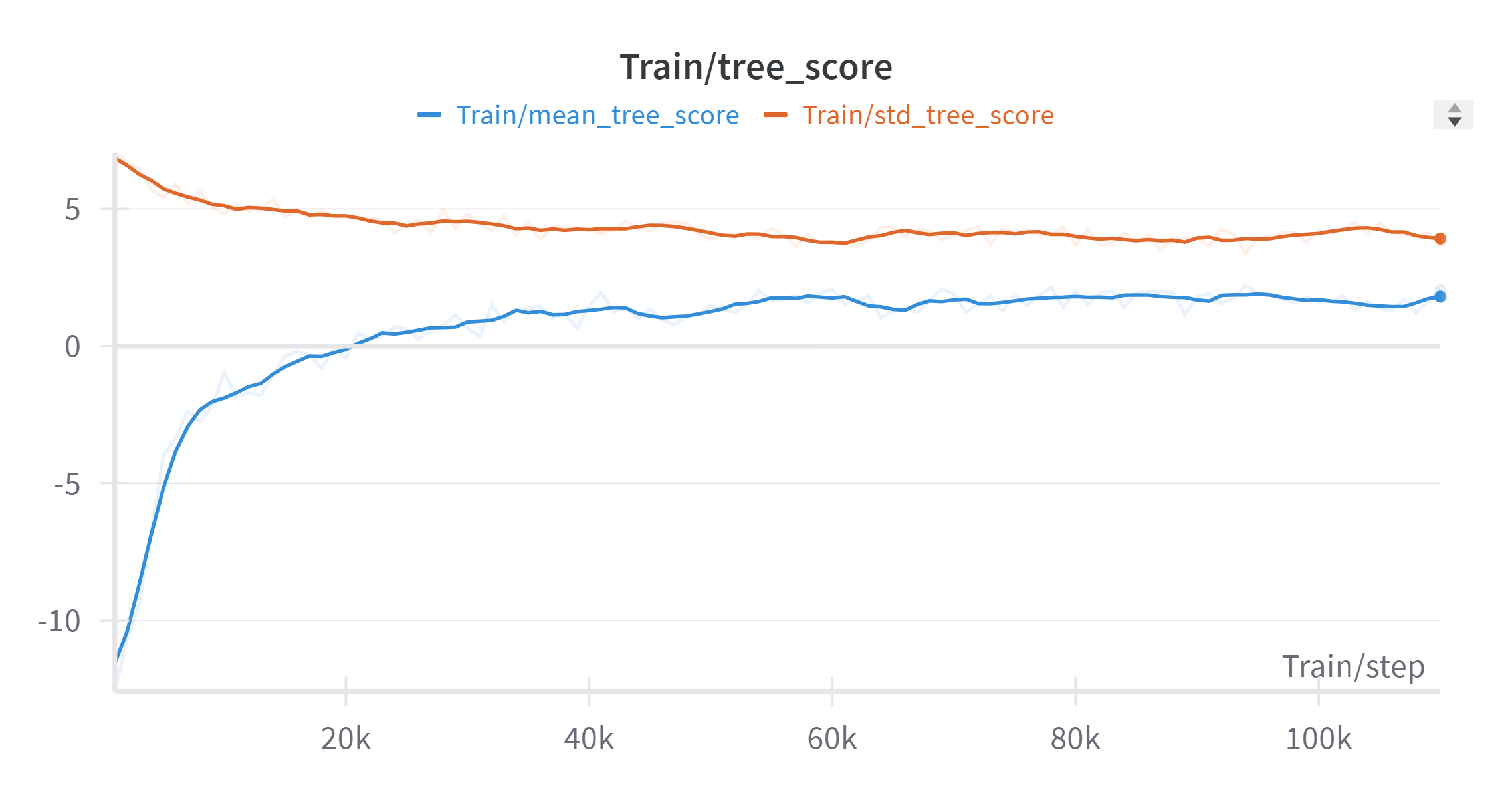}
        \caption{\small (Train): \(\mu\)=2.05 and \(\sigma\)=3.84 of `tree\_score'}
        \label{subfig:rl-fig2}
    \end{subfigure}

    \begin{subfigure}[t]{0.49\linewidth} 
        \centering
        \includegraphics[width=\linewidth]{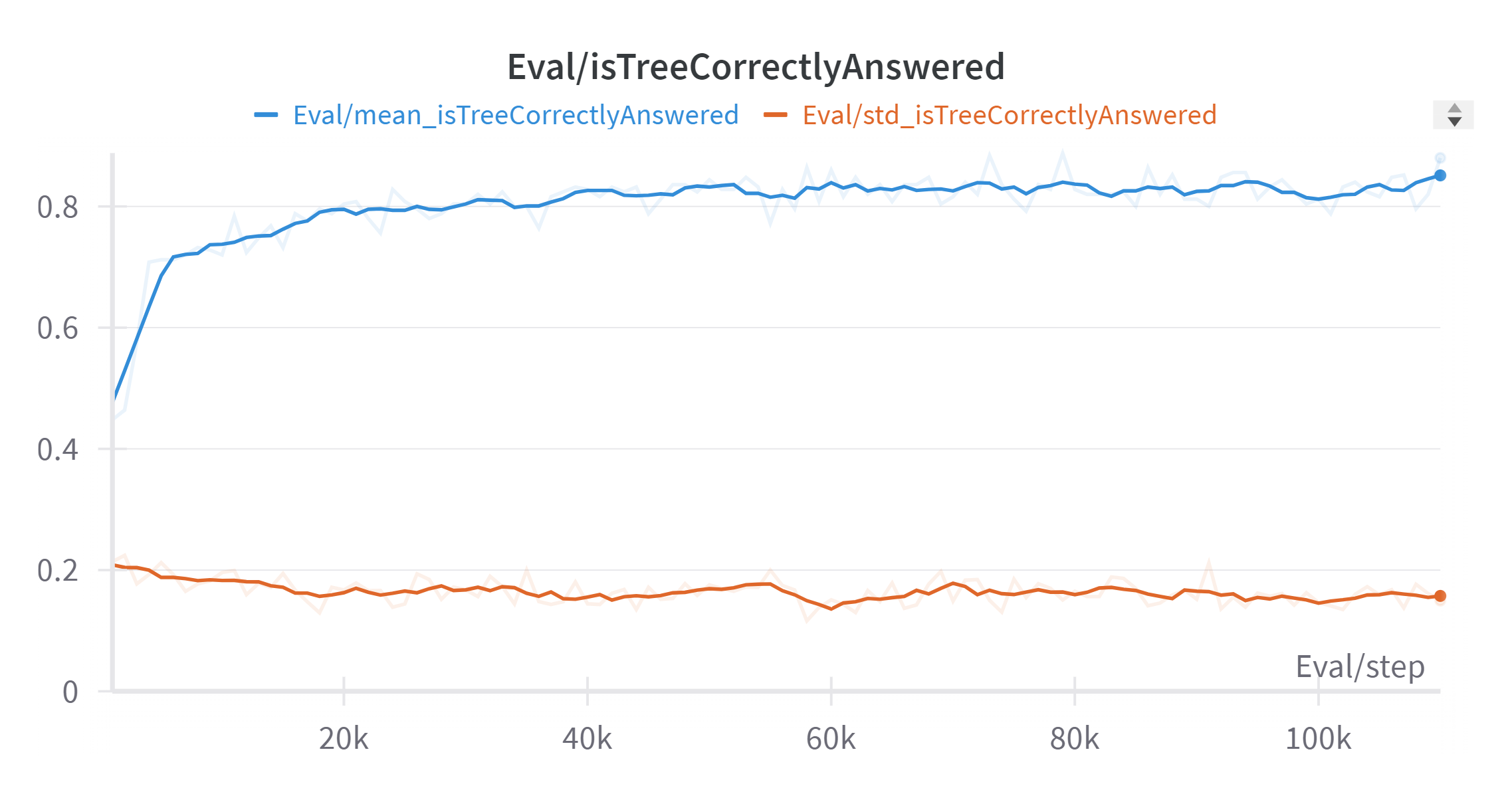}
        \caption{\small (Eval): \(\mu\)=0.88 and \(\sigma\)=0.15 of `isTreeCorrectlyAnswered'}
        \label{subfig:rl-fig3}
    \end{subfigure}
    \hfill
    \begin{subfigure}[t]{0.49\linewidth} 
        \centering
        \includegraphics[width=\linewidth]{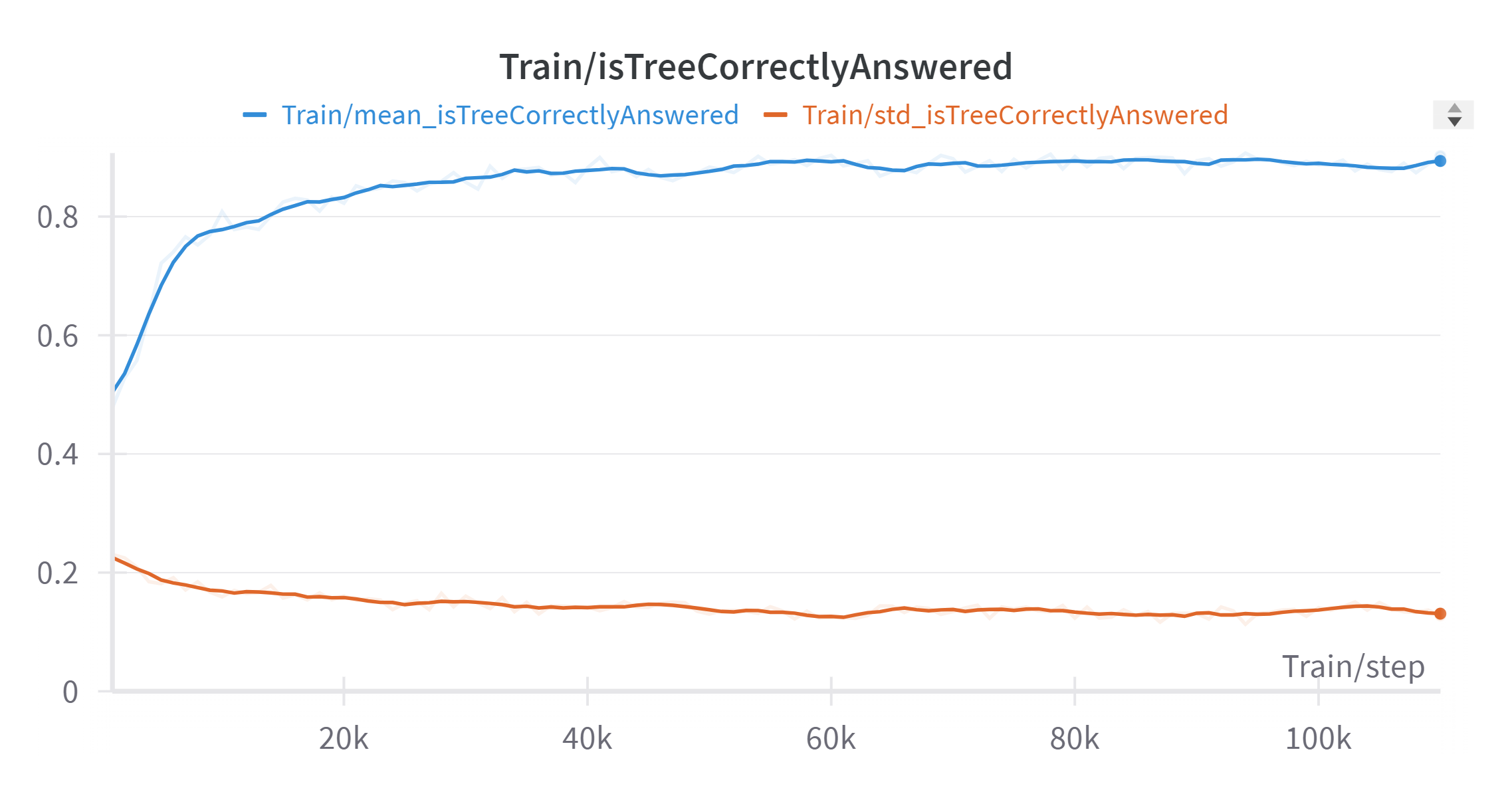}
        \caption{\small (Train): \(\mu\)=0.90 and \(\sigma\)=0.13 of `isTreeCorrectlyAnswered'}
        \label{subfig:rl-fig4}
    \end{subfigure}

    \begin{subfigure}[t]{0.49\linewidth} 
        \centering
        \includegraphics[width=\linewidth]{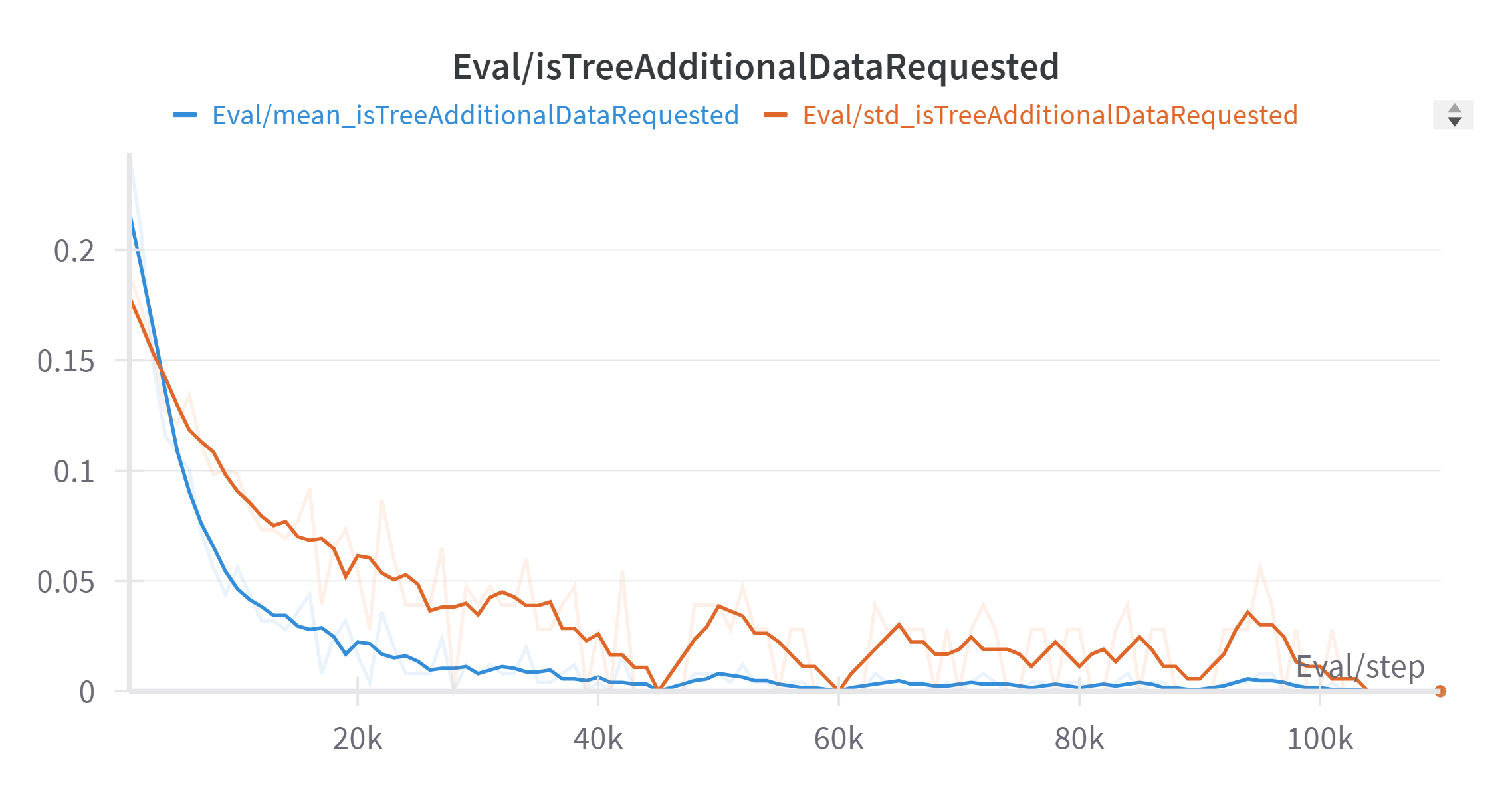}
        \caption{\small (Eval): \(\mu\)=0 and \(\sigma\)=0 of `isAdditionalDataRequested'}
        \label{subfig:rl-fig5}
    \end{subfigure}
    \hfill
    \begin{subfigure}[t]{0.49\linewidth} 
        \centering
        \includegraphics[width=\linewidth]{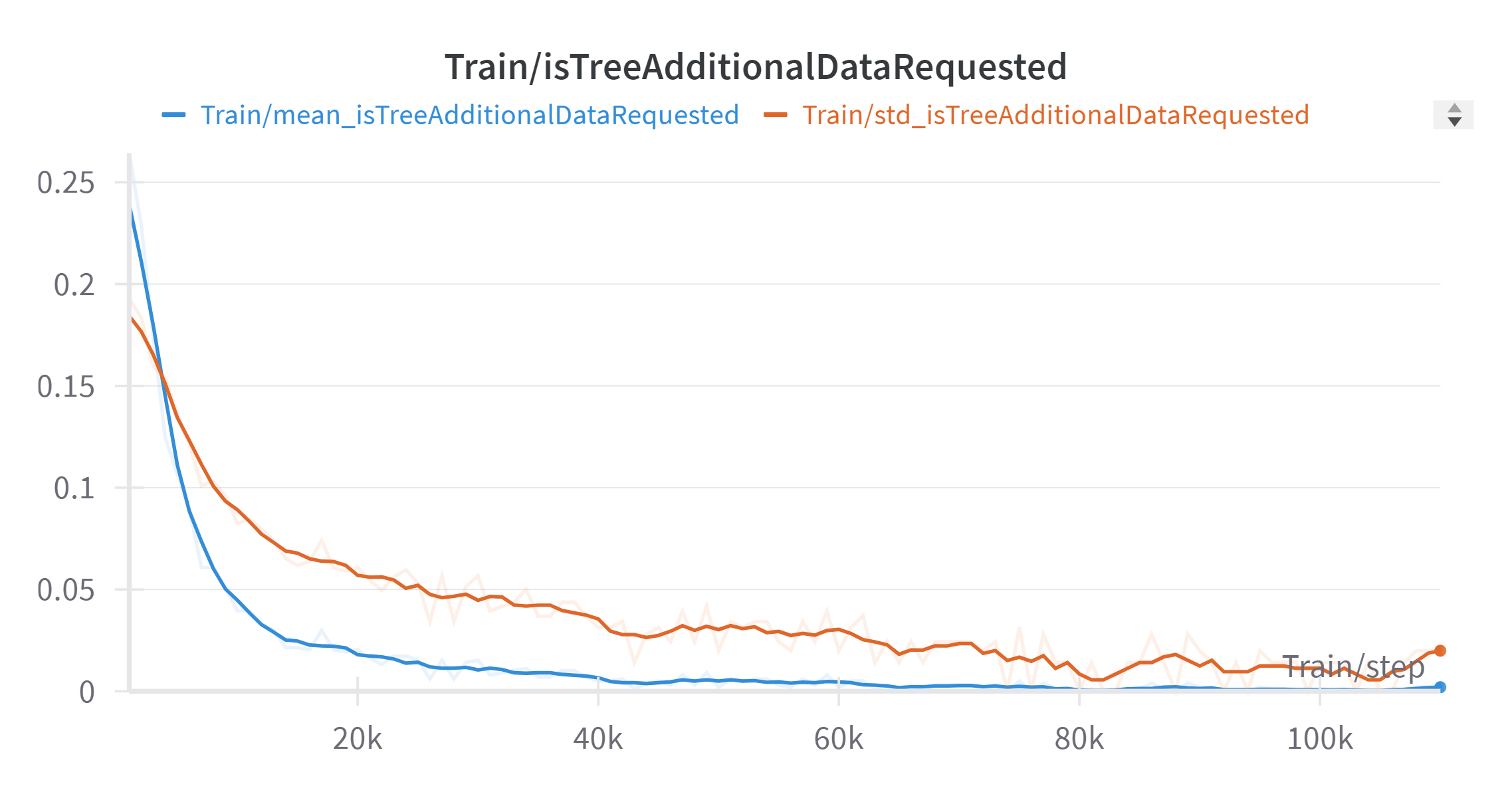}
        \caption{\small (Train): \(\mu\)=0.01 and \(\sigma\)=0 of `isAdditionalDataRequested'}
        \label{subfig:rl-fig6}
    \end{subfigure}

    \begin{subfigure}[t]{0.49\linewidth} 
        \centering
        \includegraphics[width=\linewidth]{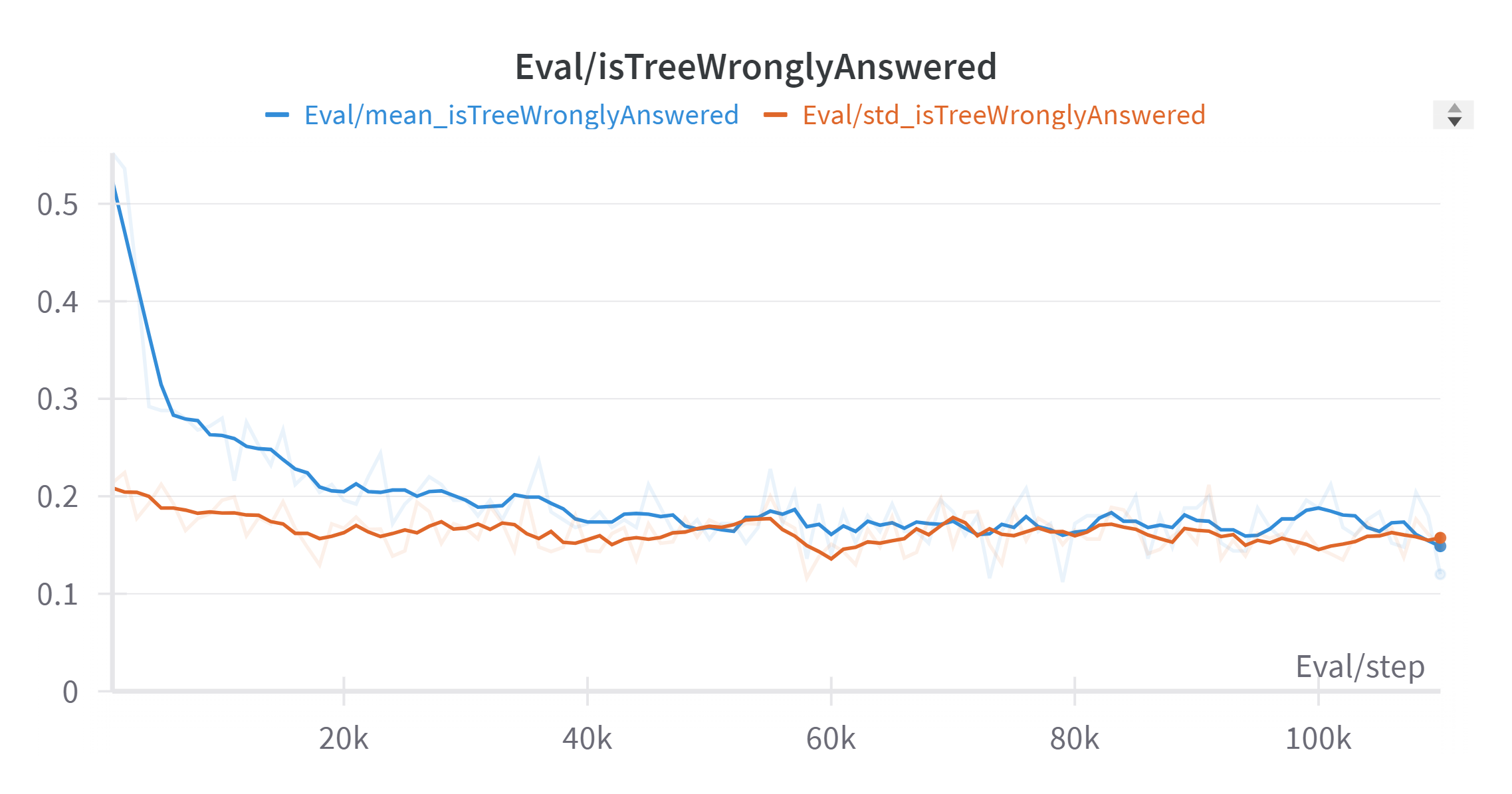}
        \caption{\small (Eval): \(\mu\)=0.12 and \(\sigma\)=0.14 of `isTreeWronglyAnswered'}
        \label{subfig:rl-fig7}
    \end{subfigure}
    \hfill
    \begin{subfigure}[t]{0.49\linewidth} 
        \centering
        \includegraphics[width=\linewidth]{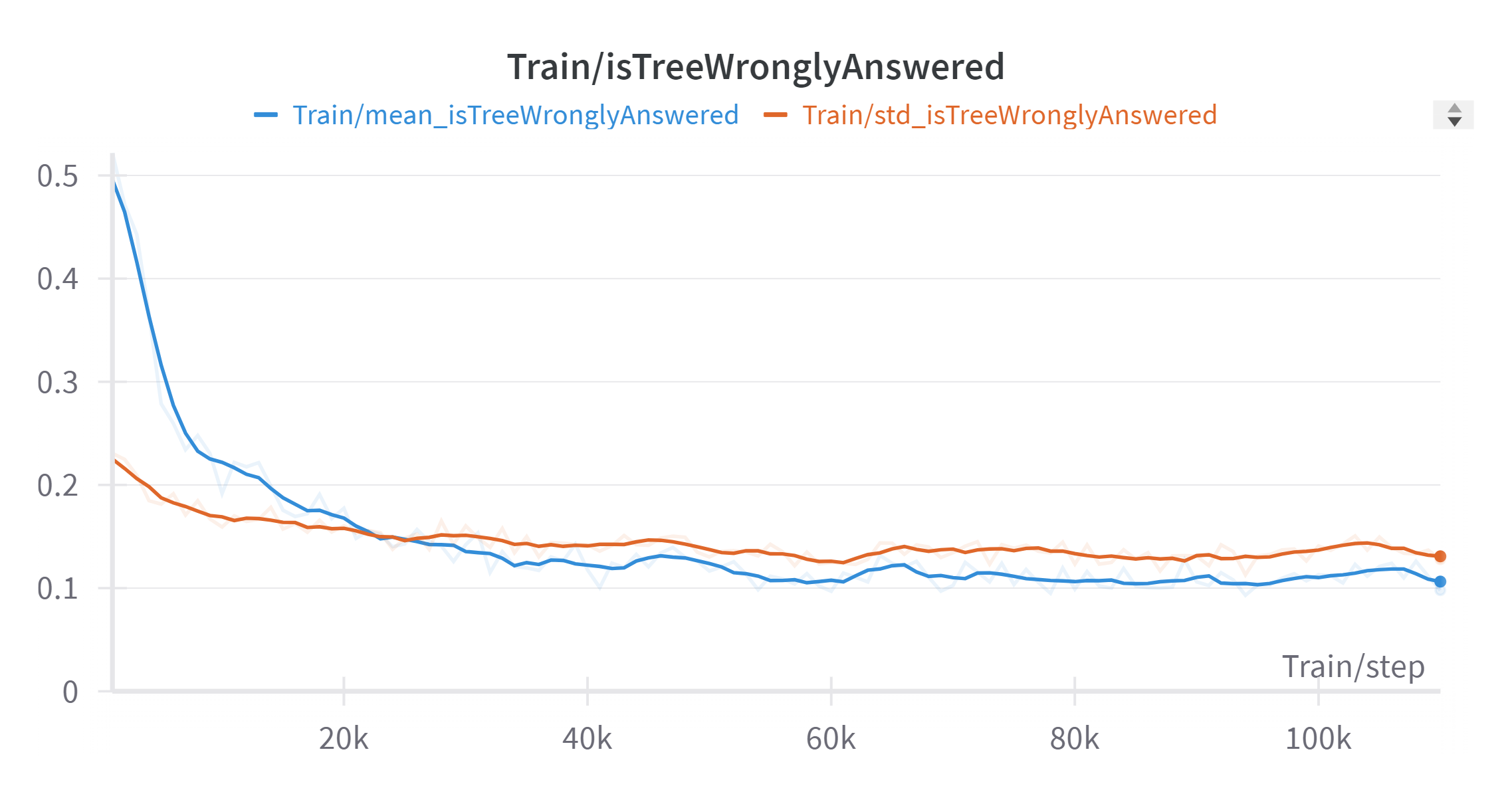}
        \caption{\small (Train): \(\mu\)=0.12 and \(\sigma\)=0.09 of `isTreeWronglyAnswered'}
        \label{subfig:rl-fig8}
    \end{subfigure}
    
    \caption{Training and evaluation metric plots for the RL agent using the A2C algorithm. (a) and (b) show the Mean and Standard Deviation of `tree\_score'. (c) and (d) display the Mean and Standard Deviation of `isTreeCorrectlyAnswered'. (e) and (f) present the Mean and Standard Deviation of `isTreeAdditionalDataRequested'. (g) and (h) show the Mean and Standard Deviation of `isTreeWronglyAnswered'.
    }
    \label{fig:rl_metric_plots}
\end{figure}

\begin{figure}[htp]
    \begin{subfigure}[t]{0.49\linewidth} 
        \centering
        \includegraphics[width=\linewidth]{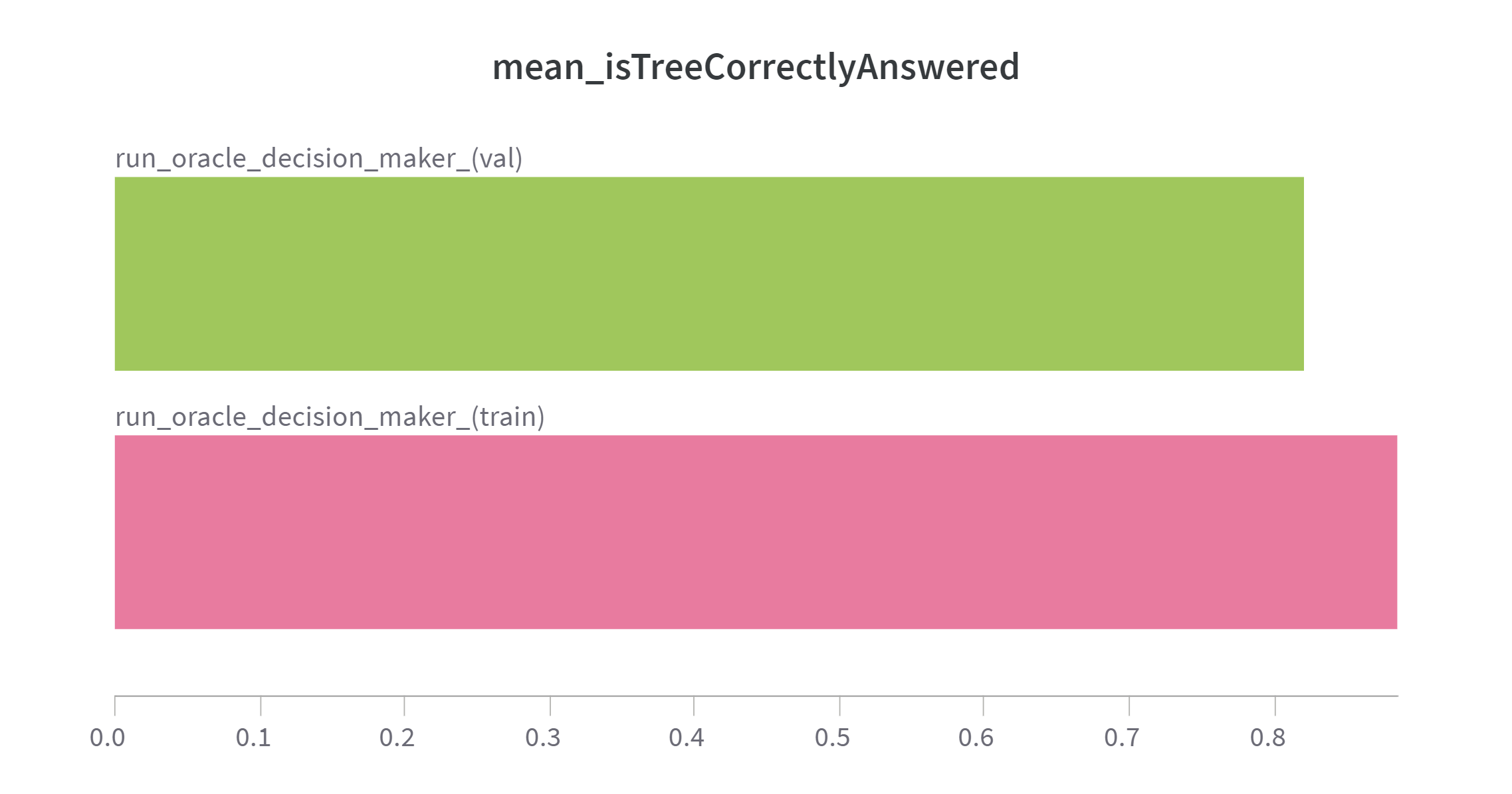}
        \caption{\small \(\mu_{\text{val}}\)=0.8201 and \(\mu_{\text{train}}\)=0.8850 for `isTreeCorrectlyAnswered'}
        \label{subfig:orc-fig1}
    \end{subfigure}
    \hfill
    \begin{subfigure}[t]{0.49\linewidth} 
        \centering
        \includegraphics[width=\linewidth]{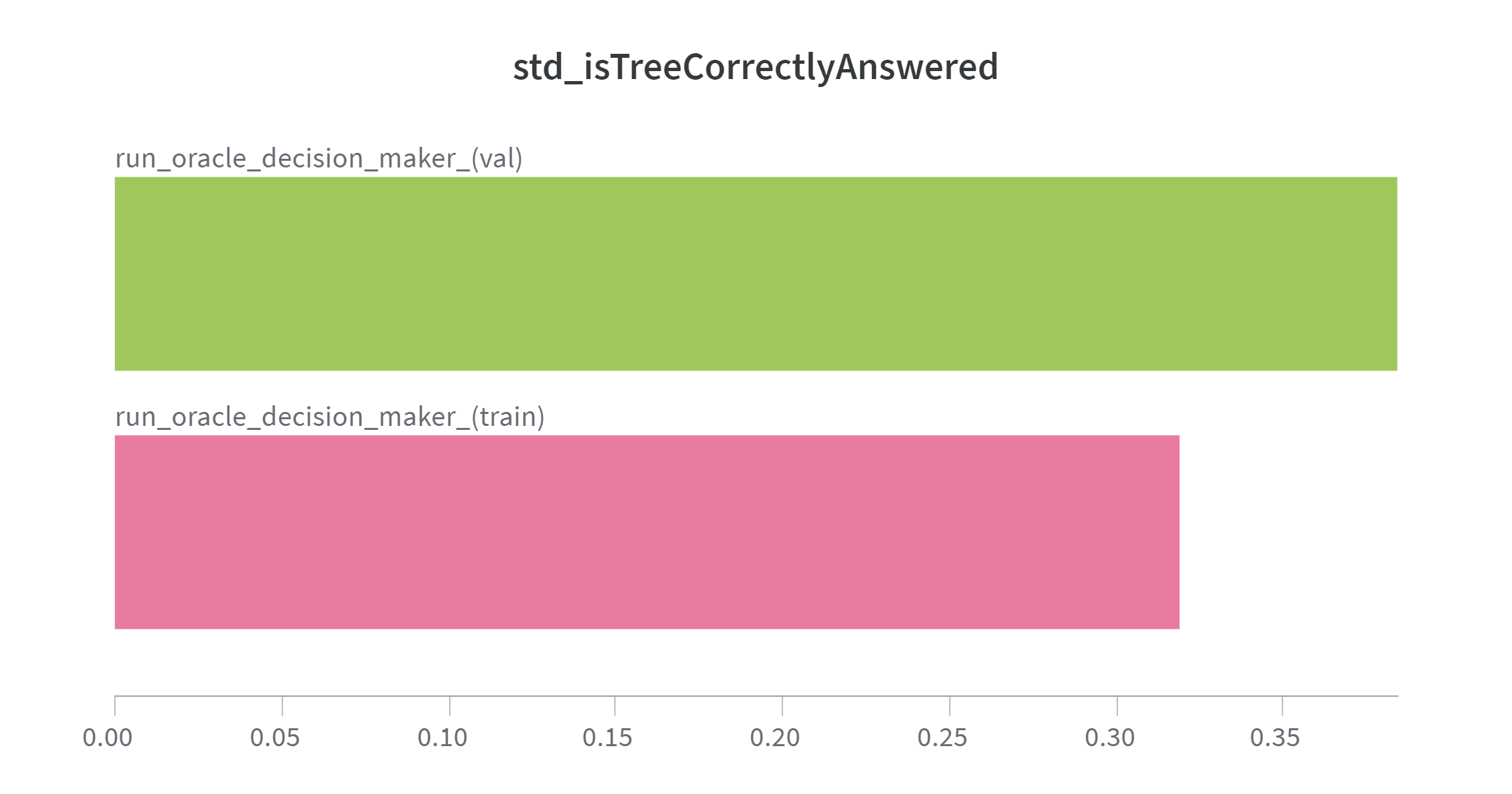}
        \caption{\small \(\sigma_{\text{val}}\)=0.3841 and \(\sigma_{\text{train}}\)=0.3191 for `isTreeCorrectlyAnswered'}
        \label{subfig:orc-fig2}
    \end{subfigure}

    \begin{subfigure}[t]{0.49\linewidth} 
        \centering
        \includegraphics[width=\linewidth]{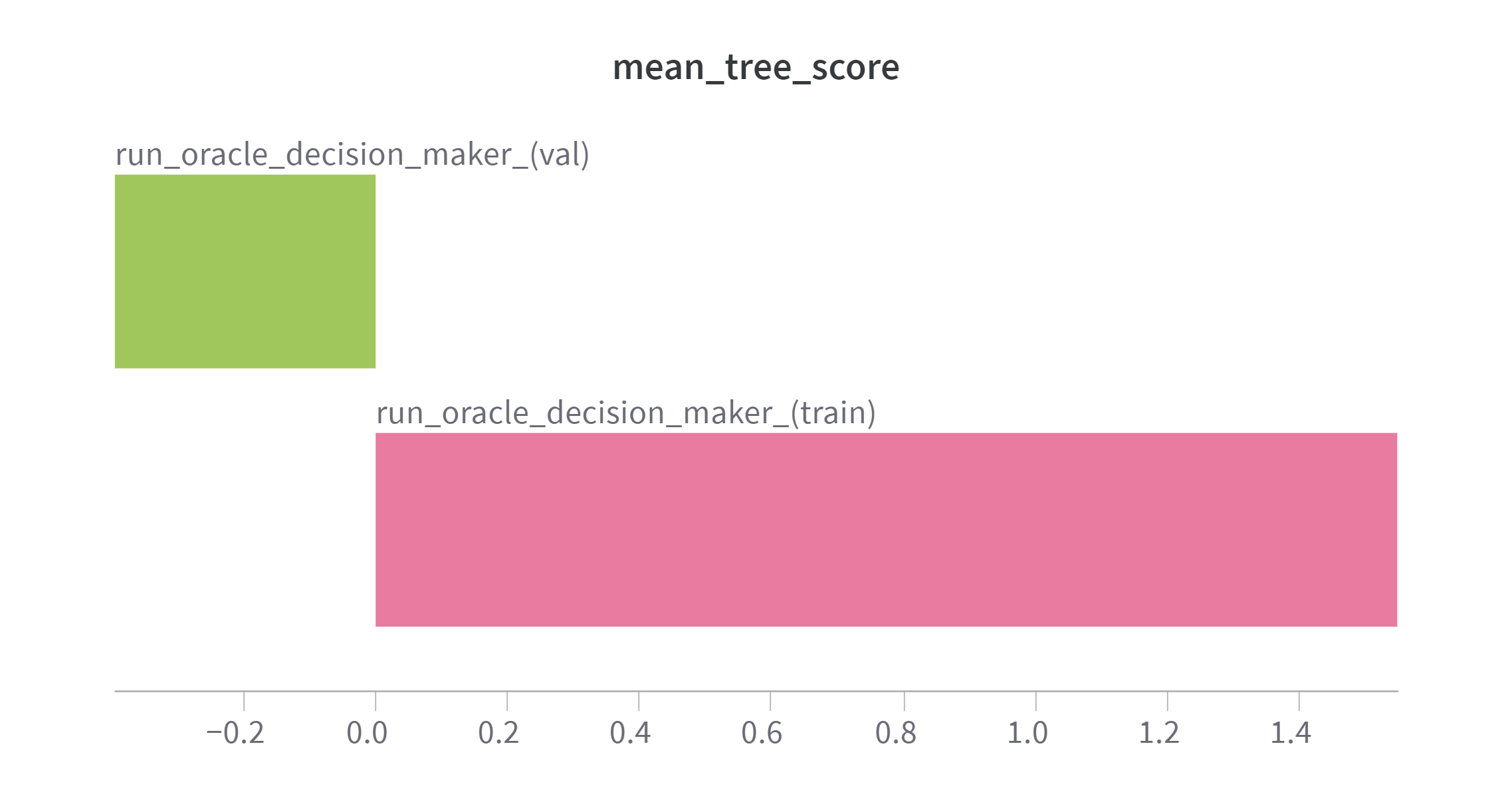}
        \caption{{\small \(\mu_{\text{val}}\)=-0.3954 and \(\mu_{\text{train}}\)=1.5491 for `tree\_score'}}
        \label{subfig:orc-fig3}
    \end{subfigure}
    \hfill
    \begin{subfigure}[t]{0.49\linewidth} 
        \centering
        \includegraphics[width=\linewidth]{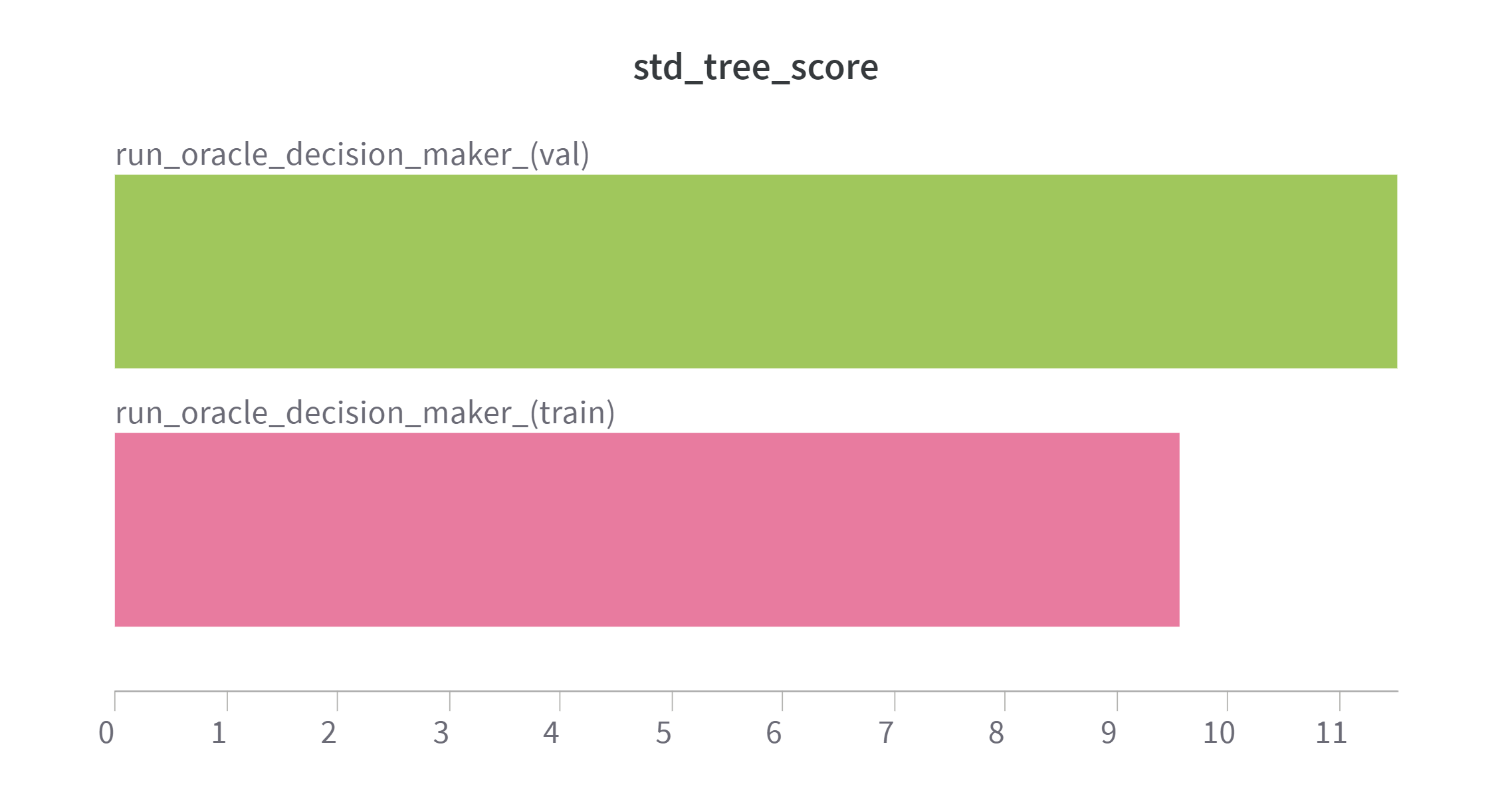}
        \caption{\small \(\sigma_{\text{val}}\)=11.5218 and \(\sigma_{\text{train}}\)=9.5717 for `tree\_score'}
        \label{subfig:orc-fig4}
    \end{subfigure}
    
    \begin{subfigure}[t]{\linewidth} 
        \centering
        \includegraphics[width=\linewidth]{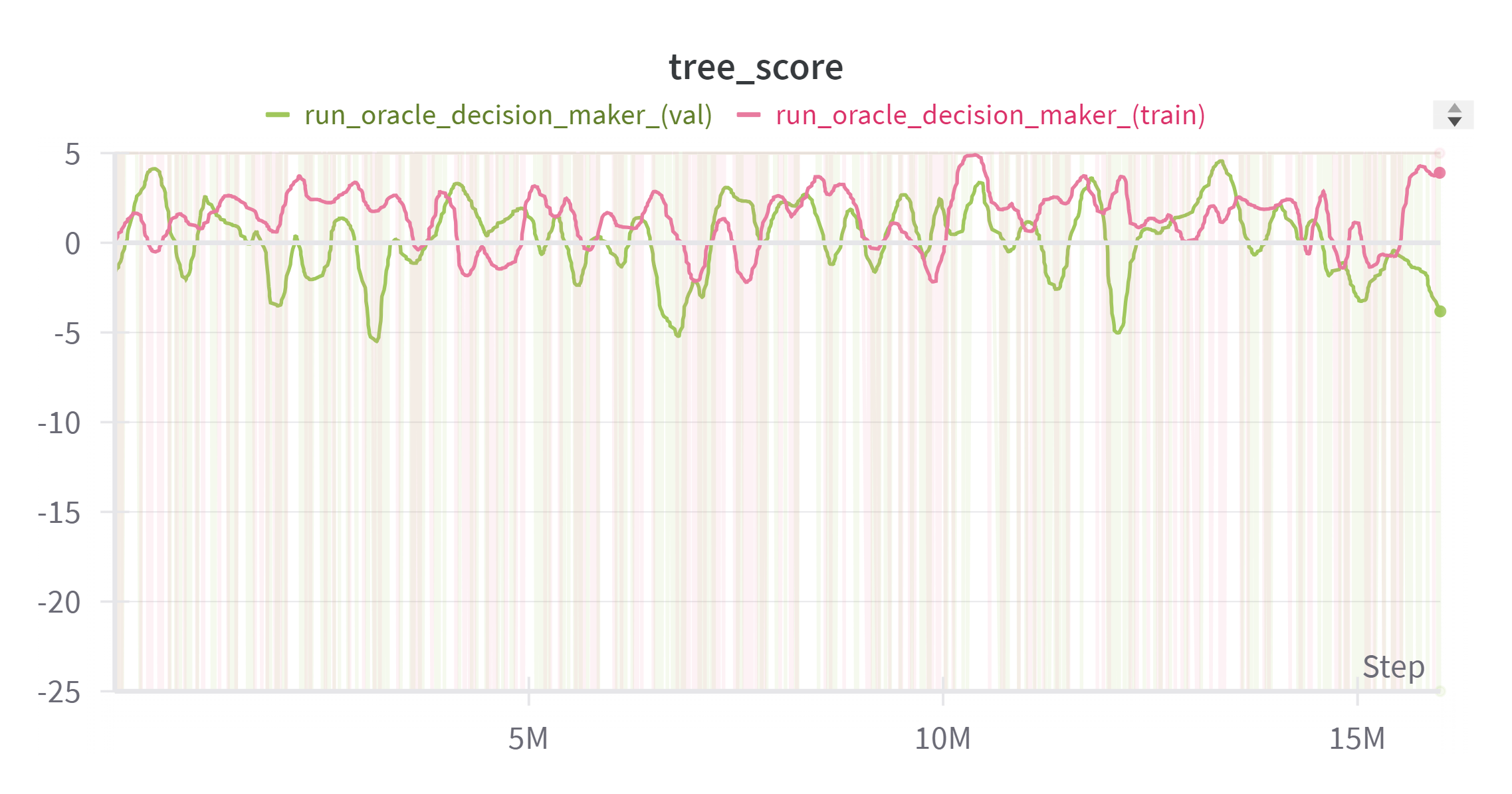}
        \caption{\small `tree\_score'}
        \label{subfig:orc-fig5}
    \end{subfigure}
    
    \caption{Training and evaluation metric plots for the Benchmark agents algorithm. (a) and (b) show the Mean and Standard Deviation of `isTreeCorrectlyAnswered'. (c) and (d) display the Mean and Standard Deviation of  `tree\_score'. (e) and (f) present the `tree\_score' logged for all scenarios with 5\% Gaussian smoothing applied to visualize the trend.
    }
    \label{fig:oracle_metrics_plot}
\end{figure}

An A2C reinforcement learning algorithm is trained across multiple \verb|Scenarios|, where each episode represents a \verb|Scenario|, and each step corresponds to a \verb|Level|. To evaluate how well the RL agent can perform and what the limit to performance is by relying solely on the outputs from the \verb|Enabler| agents (judgement insights), an \verb|Benchmark| agent is implemented. The \verb|Benchmark| serves as a benchmark agent, navigating all \verb|Scenarios| in validation/training by selecting actions based solely on the index of the maximum value in the prediction score array from the judgment data (\verb|Enabler| agent output). This benchmark is used to assess how decision-making based solely on judgment insights compares to the RL agent. Additionally, it helps define the stopping criteria for the RL agent's learning process. Performance plots for the RL agent and the \verb|Benchmark|  are shown in Figure \ref{fig:rl_metric_plots} and Figure \ref{fig:oracle_metrics_plot} respectively.

From the performance plots of the \verb|Benchmark| and the RL agent, it is seen that relying solely on judgment insights leads to uncertain decision-making, as evidenced by the standard deviation of the `tree\_score' for the \verb|Benchmark| being 11.5218 (Fig. \ref{subfig:orc-fig4})—approximately 3 times larger than that of the RL agent, which was 4.49 (Fig. \ref{subfig:rl-fig1}). 

The mean value of `isTreeCorrectlyAnswered' for the \verb|Benchmark| indicated that using only judgment insights resulted in an accuracy of 82.01\% (Fig \ref{subfig:orc-fig1}) across all validation \verb|Scenarios|, whereas the RL agent achieved 88\% accuracy (Fig \ref{subfig:rl-fig3}). The stopping criteria for the RL agent were set to trigger when the mean `isTreeCorrectlyAnswered' exceeded the \verb|Benchmark's| mean of 88\% (Fig \ref{subfig:orc-fig1}) on the training dataset and when the mean `tree\_score' surpassed the \verb|Benchmark's| mean of 1.5 (Fig \ref{subfig:orc-fig3}). According to these results, the RL agent outperformed the \verb|Benchmark| agent, with a mean accuracy 7.32\% larger than the \verb|Benchmark| agent.

The standard deviation of this metric was also strongly reduced, from 0.3841 for the \verb|Benchmark| agent to 0.15 for the RL agent, showing a 60.94\% decrease.

The training and evaluation plots of the RL agent, shown in Figure \ref{fig:rl_metric_plots}, clearly illustrate the agent's learning progression. For the RL agent, an evaluation on 1809 \verb|Scenarios| from the validation data revealed that 1736 \verb|Scenarios| achieved a perfect `tree\_score' of 5, indicating correct decisions at all \verb|Levels| for those \verb|Scenarios|. In these cases, the `isTreeWronglyAnswered' metric had a value of 0.1, meaning the agent made incorrect decisions only 10\% of the time. For \verb|Scenarios| with `tree\_scores' of 2, 3, and 4, the lower scores were due to the agent requesting additional data.

Initially, the RL agent made incorrect decisions 55.2\% of the time (Fig \ref{subfig:rl-fig7}) and requested additional data 20.44\% of the time (Fig \ref{subfig:rl-fig5}) on evaluation data. However, as training progressed, the agent significantly improved, reducing incorrect decisions to 12\%. During evaluation, the agent correctly answered all \verb|Levels| in a \verb|Scenario| 88\% of the time without requesting additional data, achieving a mean `tree\_score' of 1.4 (Fig \ref{subfig:rl-fig1}) which is 1.39 times higher than the \verb|Benchmark's| mean tree score. This demonstrates that the RL agent significantly enhances decision-making compared to relying solely on judgment data.

The summary of the most relevant results can be seen in Table \ref{tab:rl_metric_summary}.

\begin{table}[!h]
\caption{Summary of the performance of the RL and Benchmark agent showing the mean ($\mu$) and Standard Deviation ($\sigma$) for `tree\_score' and `isTreeCorrectlyAnswered' metrics.}
\label{tab:rl_metric_summary}
\centering
\begin{tabular}{|c|c|c|c|}
\hline
\textbf{Metric} & \textbf{Type} & $\mu$ & $\sigma$ \\ 
\hline
\multirow{2}{*}{\texttt{tree\_score}} & RL & 1.4 & 4.49 \\ 
& Benchmark & -0.3954 & 11.5218 \\ 
\hline
\multirow{2}{*}{\texttt{isTreeCorrectlyAnswered}} & RL & 0.88 & 0.15 \\ 
& Benchmark & 0.8201 & 0.3841 \\ 
\hline
\end{tabular}
\end{table}

\subsubsection{Human Operator}

For the human evaluation study, a total of 61 responses were recorded from participants who completed the survey on the web application, as seen in Table \ref{tab:population_statistics}. Among these, 12 were stakeholders involved in rescue or relief operations, 19 were victims affected by the disaster, and 30 were volunteers engaged in related activities. Recruiting stakeholders proved challenging due to waiting periods and their limited availability, while victims and volunteers were more willing to participate and found the survey informative. 

\begin{table}[htp]
\caption{Demographic data from the human evaluation study.}
\label{tab:population_statistics}
\centering
\begin{tabular}{lcccccc}
\hline
\textbf{Role} & \textbf{Indian} & \textbf{Japanese} & \textbf{American} & \textbf{Canadian} & \textbf{British} & \textbf{Taiwanese} \\
\hline
Stakeholder & 10 & 0 & 0 & 0 & 1 & 1\\
Volunteer & 27 & 0 & 1 & 0 & 1 & 1\\
Victim & 15 & 1 & 0 & 1 & 1 & 1 \\
\hline
\end{tabular}
\end{table}

\begin{figure}[htp]
    \begin{subfigure}[t]{0.49\linewidth} 
        \centering
        \includegraphics[width=\linewidth]{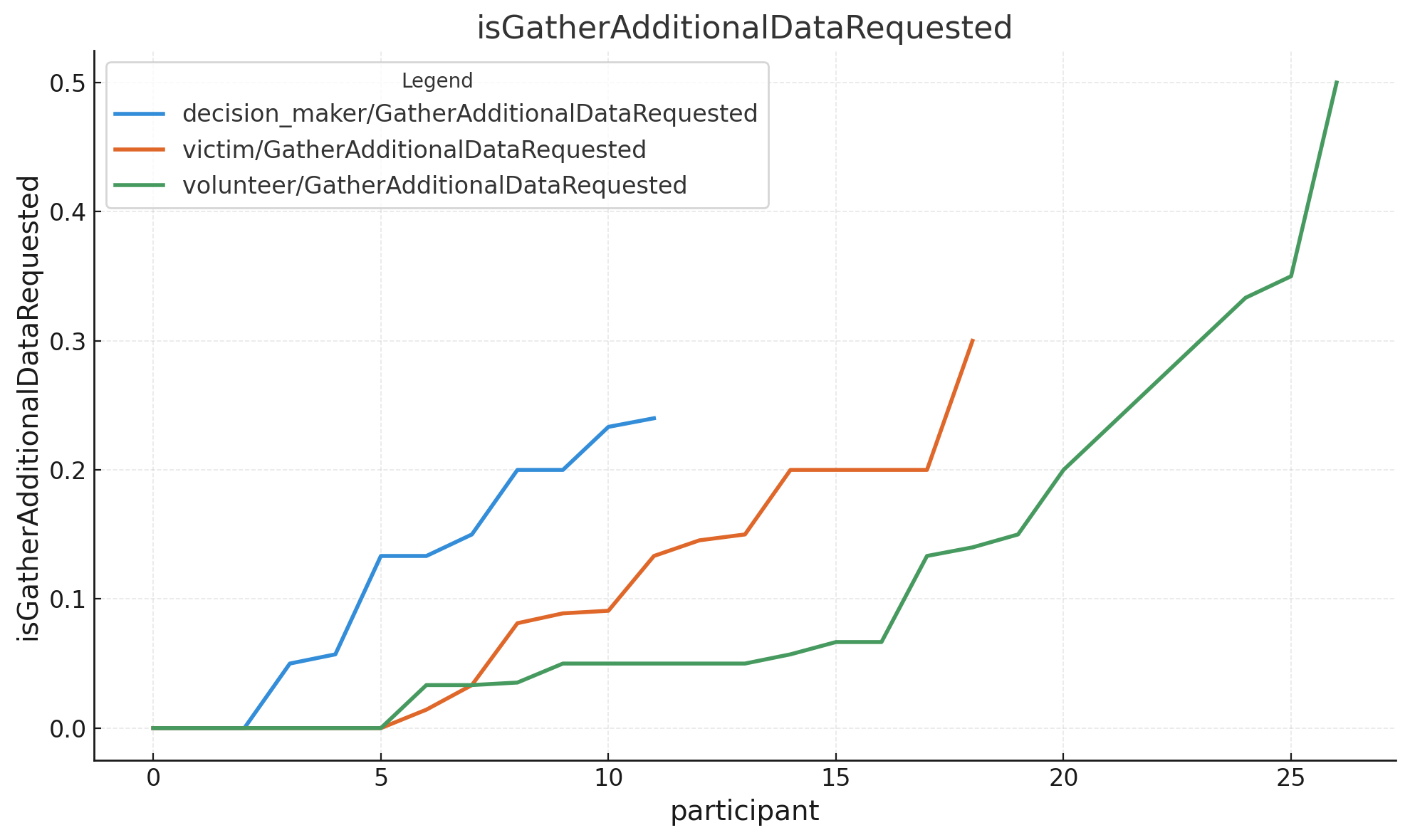}
        \caption{}
        \label{subfig:human-fig1}
    \end{subfigure}
    \hfill
    \begin{subfigure}[t]{0.49\linewidth} 
        \centering
        \includegraphics[width=\linewidth]{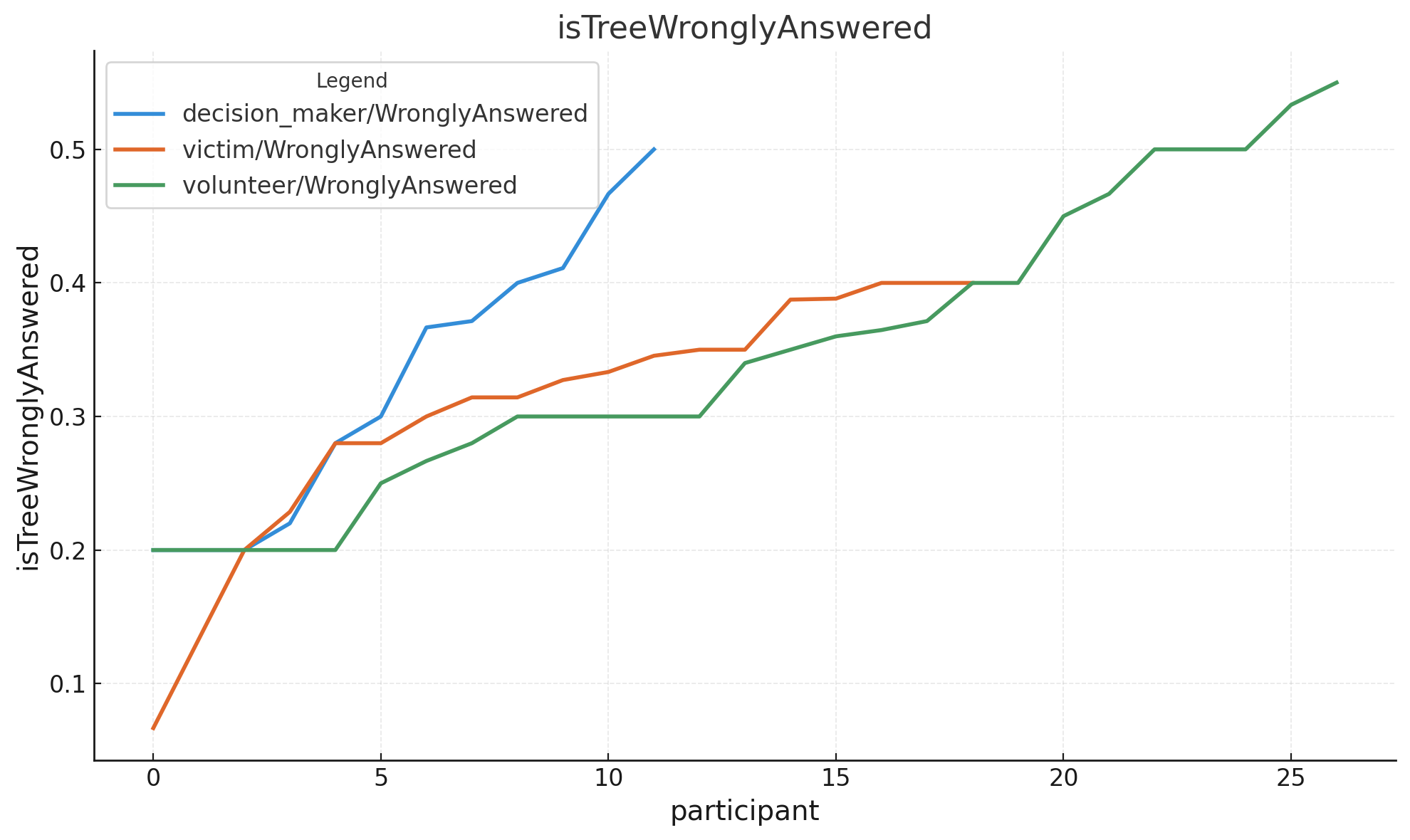}
        \caption{}
        \label{subfig:human-fig2}
    \end{subfigure}

    \begin{subfigure}[t]{0.49\linewidth} 
        \centering
        \includegraphics[width=\linewidth]{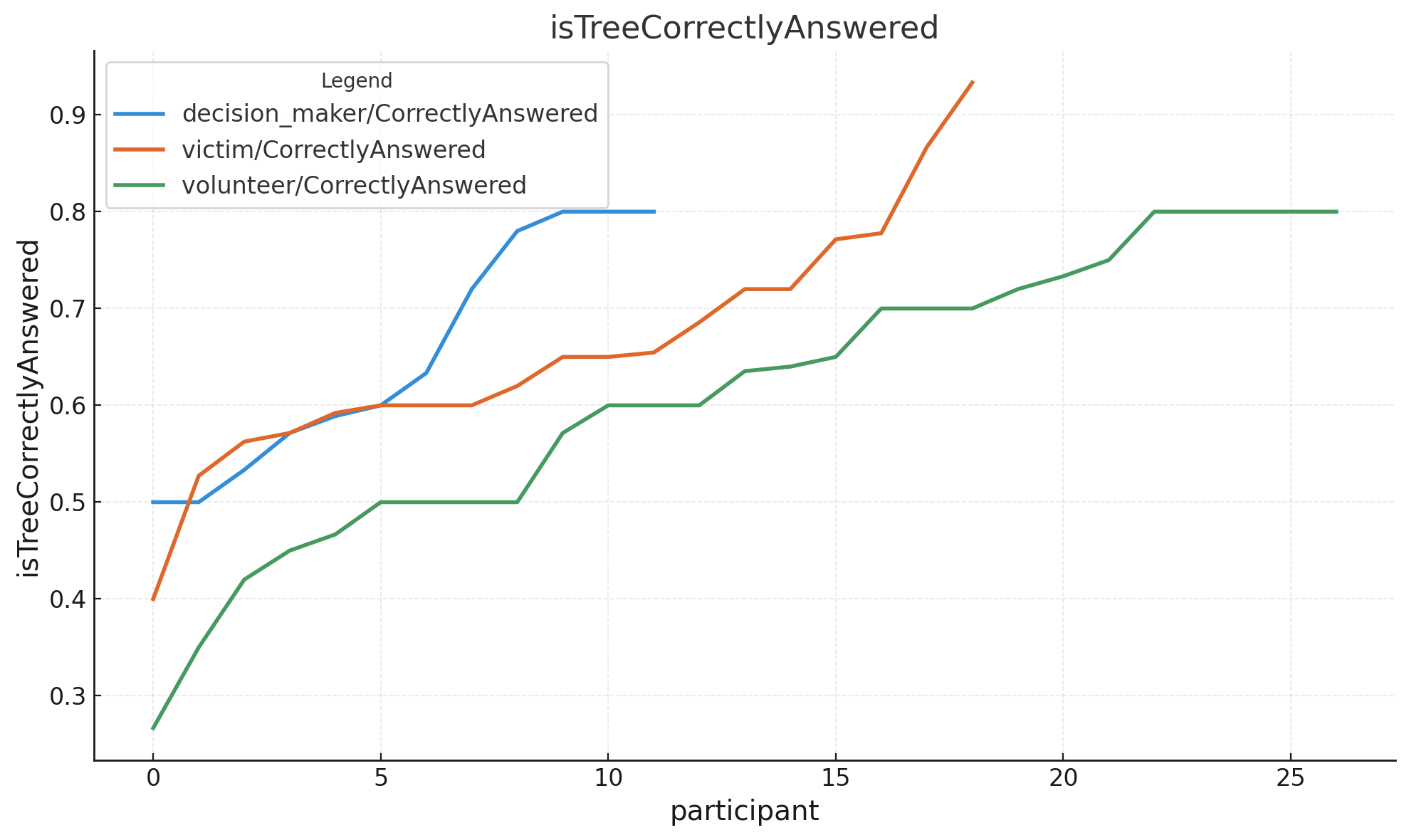}
        \caption{}
        \label{subfig:human-fig3}
    \end{subfigure}
    \hfill
    \begin{subfigure}[t]{0.49\linewidth} 
        \centering
        \includegraphics[width=\linewidth]{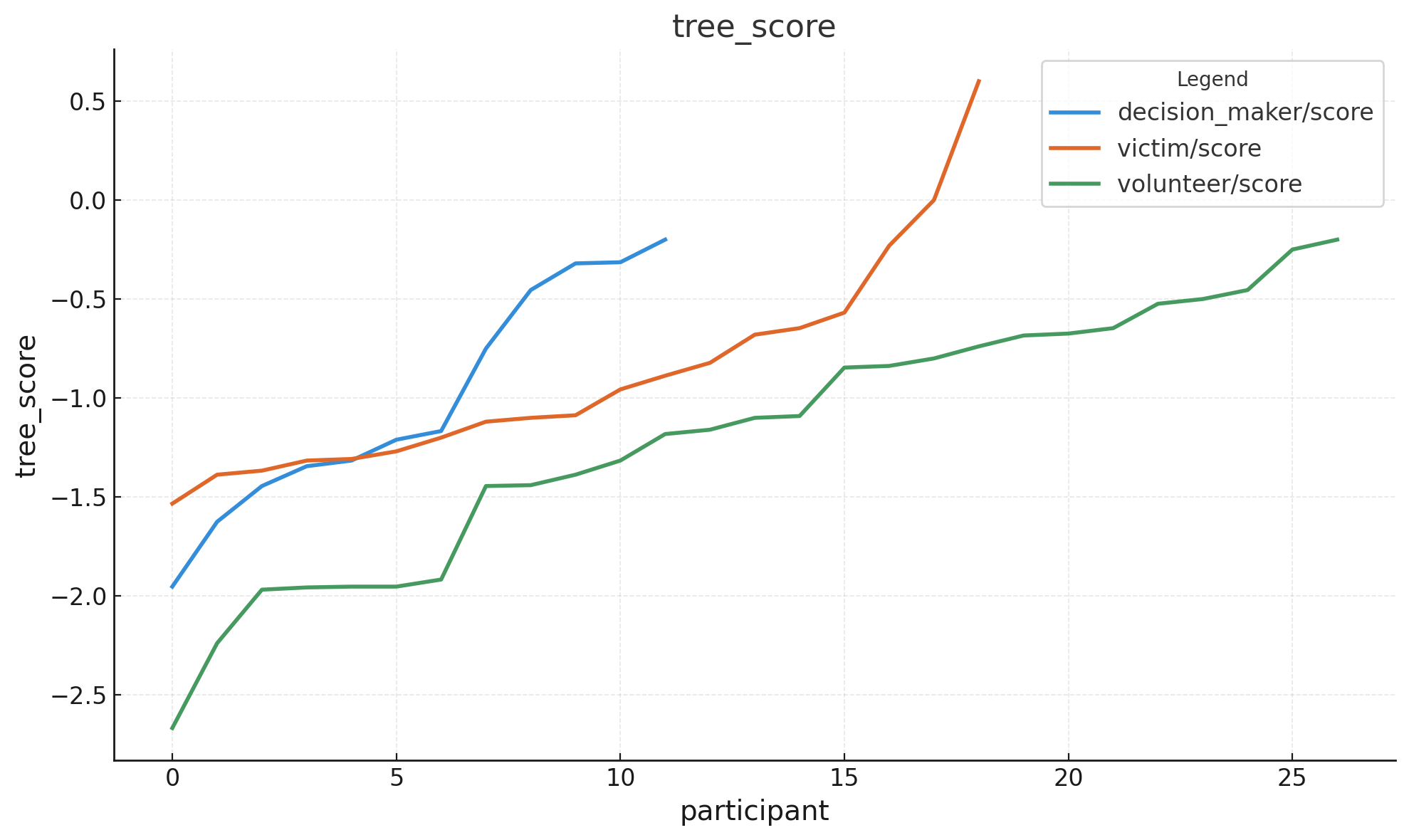}
        \caption{}
        \label{subfig:human-fig4}
    \end{subfigure}
    
    \caption{Metrics for the individual responses from the Human Operators in the survey. The horizontal axes represent each of the individual participants. (a) `isTreeAdditionalDataRequested', (b) `isTreeWronglyAnswered', (c) `isTreeCorrectlyAnswered', and (d) `tree\_score'.
    }
    \label{fig:human_metric_plots}
\end{figure}

Plots of metrics computed from individual participant responses were generated to highlight trends, and are shown in Figure \ref{fig:human_metric_plots}. 

Participants in the victim category completed an average of 10.47 \verb|Scenarios|, compared to 6.67 by stakeholders and 5.25 by volunteers. Cumulatively, victims completed 199 \verb|Scenarios|, while stakeholders and volunteers completed 80 and 142, respectively. This resulted in victims achieving a mean `tree\_score' of -0.88, which was better than the -1.08 and -1.18 scores from stakeholders and volunteers, respectively (Fig. \ref{subfig:human-fig4}). The victims' mean `tree\_score' is approximately 1.23 times better than that of stakeholders and 1.34 times better than that of volunteers. 

The difference in scores can be partly attributed to the fact that stakeholders and volunteers were more attentive to detail and less likely to make hasty decisions. This is evidenced by their higher rates of requesting additional data (Fig. \ref{subfig:human-fig1}): 11.64\% for stakeholders and 11.63\% for volunteers, compared to 9\% for victims. 

When assessing decision accuracy across the \verb|Scenarios|, all participants performed similarly: stakeholders made correct decisions 65.22\% of the time (Fig. \ref{subfig:human-fig3}) and incorrect decisions 32.63\% of the time (Fig. \ref{subfig:human-fig2}); volunteers made correct decisions 61.30\% of the time and incorrect decisions 34.75\%; and victims made correct decisions 65.80\% of the time and incorrect decisions 30.52\%. The plot reveals that, although more responses were recorded from volunteers than victims, victims achieved higher `tree\_scores'. This improvement is attributed to victims completing significantly more \verb|Scenarios| on average compared to both volunteers and stakeholders.

\subsection{Comparative Analysis}
This section presents a comparative analysis of the human evaluation study results versus the reinforcement learning (RL) decision maker. 

\subsubsection{RL vs Stakeholder}
\label{subsubsec:rl_vs_stakeholder}
This section provides a comparative analysis of the aggregate performance of all stakeholders and the reinforcement learning agent. Additionally, comparisons are made with the stakeholder who completed the most \verb|Scenarios|, against the RL agent.

From Table \ref{tab:stakeholder_vs_rl_agg}, it is evident that the reinforcement learning (RL) agent outperforms the aggregate of stakeholder responses across all metrics. The RL agent achieves a Mean Tree Score (M.T.S) of 1.4, significantly higher than the stakeholder's score of -1.0082. In terms of accuracy, the RL agent makes correct decisions 88\% of the time (M.C.A = 0.88), whereas stakeholders collectively achieve a lower accuracy of 65.22\% (M.C.A = 0.6522). Furthermore, the RL agent makes fewer incorrect decisions, with a Mean `isTreeWronglyAnswered' (M.W.A) of 12\%, which is approximately 0.37 times smaller than the stakeholder's M.W.A of 32.63\%. Finally, the RL agent does not request additional data at all (M.A.D = 0), in contrast to stakeholders, who request additional data 11.64\% of the time (M.A.D = 0.1164). Overall, the RL agent demonstrates superior performance in decision-making, accuracy, and efficiency compared to the collective stakeholder responses.

\begin{table}[htp]
\caption{Results comparing the RL agent with the Stakeholder participant responses: Stakeholder A (Aggregate of all stakeholder responses), and Stakeholder B (Most Scenarios Completed), showing: Mean Tree Score (M.T.S), Mean `isTreeCorrectlyAnswered' (M.C.A), Mean `isTreeWronglyAnswered' (M.W.A), and Mean `isGatherAdditionalDataRequested' (M.A.D).}
\label{tab:stakeholder_vs_rl_agg}
\centering
\begin{tabular}{lcccc}
\hline
\textbf{Agent} & \textbf{M.T.S} & \textbf{M.C.A} & \textbf{M.W.A} & \textbf{M.A.D} \\
\hline
Stakeholder A (Collective) & -1.0082 & 0.6522 & 0.3263 & 0.1164 \\
Stakeholder B (Most Scenarios) & -1.3443 & 0.5889 & 0.4111 & 0.1333 \\
\textbf{RL Agent} & \textbf{1.4000} & \textbf{0.8800} & \textbf{0.1200} & \textbf{0.0000} \\
\hline
\end{tabular}
\end{table}

Additionally, from Table \ref{tab:stakeholder_vs_rl_agg}, an analysis between the RL agent and Stakeholder B (the stakeholder participant who completed 18 \verb|Scenarios|) reveals that stakeholder tends to make less accurate decisions as they complete more \verb|Scenarios|, only 58.89\% accuracy, across the 18 \verb|Scenarios| they completed, whereas the RL agent achieves 88\% accuracy across the validation data, which is 1809 \verb|Scenarios|. The observed trend of stakeholders making more inaccurate decisions as they complete additional \verb|Scenarios| can be attributed to community disaster fatigue, particularly the sense of defeatism. As highlighted in \cite{INGHAM2023103831}, prolonged exposure to disaster-related decision-making often results in decreased accuracy and effectiveness among stakeholders.

\subsubsection{RL vs Volunteer}
This section provides a comparative analysis between the aggregate performance of all volunteers and the reinforcement learning agent. Additionally, comparisons are made with the volunteer participant who completed the most \verb|Scenarios|, against the RL agent. 

From Table \ref{tab:volunteer_vs_rl_agg}, it is evident that the reinforcement learning (RL) agent outperforms the aggregate of volunteer responses across all metrics. The RL agent achieves a Mean Tree Score (M.T.S) of 1.4, significantly higher than the volunteer's score of -1.1825. In terms of accuracy, the RL agent makes correct decisions 88\% of the time (M.C.A = 0.88), whereas volunteers collectively achieve a lower accuracy of 61.31\% (M.C.A = 0.6131). Furthermore, the RL agent makes fewer incorrect decisions, with a Mean `isTreeWronglyAnswered' (M.W.A) of 12\%, which is approximately 0.35 times smaller than the volunteers' M.W.A of 34.75\%. Finally, the RL agent does not request additional data at all (M.A.D = 0), in contrast to volunteers, who request additional data 11.66\% of the time (M.A.D = 0.1166). Overall, the RL agent demonstrates superior performance in decision-making, accuracy, and efficiency compared to the collective volunteer responses.

\begin{table}[htp]
\caption{Results comparing the RL agent with the Volunteer participant responses: Volunteer A (Aggregate of all stakeholder responses), and Volunteer B (Most Scenarios Completed), showing: Mean Tree Score (M.T.S), Mean `isTreeCorrectlyAnswered' (M.C.A), Mean `isTreeWronglyAnswered' (M.W.A), and Mean `isGatherAdditionalDataRequested' (M.A.D).}
\label{tab:volunteer_vs_rl_agg}
\centering
\begin{tabular}{lcccc}
\hline
\textbf{Agent} & \textbf{M.T.S} & \textbf{M.C.A} & \textbf{M.W.A} & \textbf{M.A.D} \\
\hline
Volunteer A (Collective) & -1.1825 & 0.6131 & 0.3475 & 0.1166 \\
Volunteer B (Most Scenarios) & -1.1818 & 0.6353 & 0.3647 & 0.0353  \\
\textbf{RL Agent} & \textbf{1.4000} & \textbf{0.8800} & \textbf{0.1200} & \textbf{0.0000} \\
\hline
\end{tabular}
\end{table}

Additionally, from Table \ref{tab:volunteer_vs_rl_agg}, an analysis between the RL agent and Volunteer B (the volunteer participant who completed 17 \verb|Scenarios|) reveals that volunteers tend to make less accurate decisions as they complete more \verb|Scenarios|, only 63.53\% accuracy, across the 17 \verb|Scenarios| they completed, whereas the RL agent achieves 88\% accuracy across the validation data, which is 1809 \verb|Scenarios|. The observed trend of volunteers making more inaccurate decisions as they complete additional \verb|Scenarios| can also be attributed to community disaster fatigue, similar to the observed trend seen for stakeholder responses in Section:\ref{subsubsec:rl_vs_stakeholder}.

\subsubsection{RL vs Victim}
This section offers a comparative analysis between the aggregate performance of all victim participants and the reinforcement learning (RL) agent. Additionally, it compares the RL agent with the victim participant who completed the most \verb|Scenarios|. Although victims typically do not participate in decision-making processes within disaster management, this analysis provides insights into how individuals from different backgrounds perform against an RL agent in disaster-related decision-making. The comparison is based on the learning curve of the victims in understanding disaster-related decision-making using the information and tools provided to the victim participants through the web application.

From Table \ref{tab:victim_vs_rl_agg}, it is evident that the reinforcement learning (RL) agent outperforms the aggregate of victim responses across all metrics. The RL agent achieves a Mean Tree Score (M.T.S) of 1.4, significantly higher than the victim's score of -0.8835. In terms of accuracy, the RL agent makes correct decisions 88\% of the time (M.C.A = 0.88), whereas victims collectively achieve a lower accuracy of 65.80\% (M.C.A = 0.6580). Furthermore, the RL agent makes fewer incorrect decisions, with a Mean `isTreeWronglyAnswered' (M.W.A) of 12\%, which is approximately 0.39 times smaller than the victims's M.W.A of 30.52\%. Finally, the RL agent does not request additional data at all (M.A.D = 0), in contrast to victims, who request additional data 9.67\% of the time (M.A.D = 0.0967). Overall, the RL agent demonstrates superior performance in decision-making, accuracy, and efficiency compared to the collective volunteer responses.

\begin{table}[htp]
\caption{Results comparing the RL agent with the Victim participant responses: Victim A (Aggregate of all stakeholder responses), and Victim B (Most Scenarios Completed), showing: Mean Tree Score (M.T.S), Mean `isTreeCorrectlyAnswered' (M.C.A), Mean `isTreeWronglyAnswered' (M.W.A), and Mean `isGatherAdditionalDataRequested' (M.A.D).}
\label{tab:victim_vs_rl_agg}
\centering
\begin{tabular}{lcccc}
\hline
\textbf{Agent} & \textbf{M.T.S} & \textbf{M.C.A} & \textbf{M.W.A} & \textbf{M.A.D} \\
\hline
Victim A (Collective) & -0.8835 & 0.6580 & 0.3052 & 0.0967 \\
Victim B (Most Scenarios) & -1.3077 & 0.5922 & 0.3882 & 0.0000  \\
\textbf{RL Agent} & \textbf{1.4000} & \textbf{0.8800} & \textbf{0.1200} & \textbf{0.0000} \\
\hline
\end{tabular}
\end{table}

Additionally, from Table \ref{tab:victim_vs_rl_agg}, an analysis between the RL agent and Victim B (the victim participant who completed 51 \verb|Scenarios|) reveals that victims tend to make less accurate decisions as they complete more \verb|Scenarios|, only 59.22\% accuracy which is almost random decisions, across the 51 \verb|Scenarios| they completed, whereas the RL agent achieves 88\% accuracy across the validation data, which is 1809 \verb|Scenarios|. Participants identified as victims tended to complete more \verb|Scenarios|, but this did not translate into more accurate decision-making. Their decisions appeared almost random across the \verb|Scenarios|.

\subsubsection{RL vs All}
This section offers a comparative analysis between the aggregate performance of all participants, independent of their role, and the reinforcement learning (RL) agent.

From Table \ref{tab:all_vs_rl_agg}, it is evident that the reinforcement learning (RL) agent outperforms the aggregate of all responses across all metrics. The RL agent achieves a Mean Tree Score (M.T.S) of 1.4, significantly higher than the collective participant score of -1.0651. In terms of accuracy, the RL agent makes correct decisions 88\% of the time (M.C.A = 0.88), whereas collectively the participants achieve a lower accuracy of 63.34\% (M.C.A = 0.6334). According to these results, the RL agent outperformed the participants, with a mean accuracy 38.93\% larger than the participants.

Furthermore, the RL agent makes fewer incorrect decisions, with a Mean `isTreeWronglyAnswered' (M.W.A) of 12\%, which is approximately 2.75 times smaller than the collective of all the participants' M.W.A (33.01\%). Finally, the RL agent does not request additional data at all (M.A.D = 0), in contrast, the participants collectively request additional data 10.42\% of the time (M.A.D = 0.1042). Overall, the RL agent demonstrates superior performance in decision-making, accuracy, and efficiency compared to the collective volunteer responses.

\begin{table}[htp]
\caption{Results comparing the RL agent with the all participant responses showing: Mean Tree Score (M.T.S), Mean `isTreeCorrectlyAnswered' (M.C.A), Mean `isTreeWronglyAnswered' (M.W.A), and Mean `isGatherAdditionalDataRequested' (M.A.D).}
\label{tab:all_vs_rl_agg}
\centering
\begin{tabular}{lcccc}
\hline
\textbf{Agent} & \textbf{M.T.S} & \textbf{M.C.A} & \textbf{M.W.A} & \textbf{M.A.D} \\
\hline
All (Collective) & -1.0651 & 0.6334 & 0.3301 & 0.1042 \\
\textbf{RL Agent} & \textbf{1.4000} & \textbf{0.8800} & \textbf{0.1200} & \textbf{0.0000} \\
\hline
\end{tabular}
\end{table}

\subsection{Discussions Summary}
From the results for the \verb|Decision Maker| agents, it is quite evident that the proposed framework for structured decision-making, has enabled the reinforcement learning \verb|Decision| \verb|Maker| to outperform the \verb|Benchmark| agent and the Human operators, with a mean `tree\_score' 1.39 times larger than the \verb|Benchmark| agent and 1.62 times larger than the collective mean `tree\_score' score of the human participants from the study, irrespective of their role. In addition, the RL agent outperformed the \verb|Benchmark| agent and the Human operators, with a mean accuracy (`isTreeCorrectlyAnswered') 7.32\% larger than the \verb|Benchmark| agent and 38.93\% larger than the accuracy of decisions made across multiple scenarios by the Human operators. Also, the stability of the RL agent making consistently correct decisions is 60.94\% higher than the \verb|Benchmark| agent, which is calculated as the reduction in the standard deviations of the accuracy.

The stability in decision-making provided by this framework is critical for the safety-critical decisions required in disaster management. The performance statistics clearly demonstrate that integrating structure into autonomous decision-making can effectively make decisions more reliable and justifiable, mitigating challenges such as community fatigue, inter-agency coordination issues, and data overload faced by stakeholders. To further enhance the suitability of this framework for practical applications, incorporating validation checks for ethical and legal compliance, along with safety mechanisms for irreversible decisions, is essential.


\section{Conclusions}\label{sec6}
This paper proposed a structured decision-making framework for autonomous decision-making in the context of disaster management. The framework introduced concepts of \verb|Scenarios|, \verb|Levels| and \verb|Enabler| agents that aid the reinforcement learning \verb|Decision Maker| agent in autonomous decision-making. The framework's performance was rigorously evaluated against systems that rely solely on judgement-based insights, \verb|Benchmark| agents, as well as human operators who have disaster experience: victims, volunteers, and stakeholders. The results demonstrate that the structured decision-making framework achieves 60.94\% greater stability in consistently accurate decisions across multiple \verb|Scenarios|, compared to judgement-based systems. Moreover, the framework was shown to outperform human operators with a 38.93\% higher accuracy across various \verb|Scenarios|. This research highlights the potential benefits of integrating structure into safety-critical decision-making, paving the way for future improvements. 


\section{Data Availability}
The code used to showcase the experiments carried out in the research is publicly available at \url{https://github.com/From-Governance-To-Autonomous-Robots/Autonomous-Governance-in-Disaster-Management}.

\bibliography{sn-bibliography}
\end{document}